\documentclass[11pt]{article}

\usepackage[utf8]{inputenc}
\usepackage{lmodern}
\usepackage[T1]{fontenc}
\usepackage[margin=1in]{geometry}
\usepackage{microtype}
\usepackage{amsmath,amssymb,amsthm}

\theoremstyle{definition}
\newtheorem{definition}{Definition}
\usepackage{booktabs}
\usepackage{tabularx}
\usepackage{tikz}
\usepackage{pgfplots}
\pgfplotsset{compat=1.17}
\usetikzlibrary{arrows.meta,positioning,shapes.geometric}
\usepackage{csquotes}
\MakeOuterQuote{"}
\usepackage[colorlinks=true,citecolor=blue,linkcolor=blue,urlcolor=blue]{hyperref}

\usepackage[style=authoryear,backend=biber,natbib=true,giveninits=true,
            uniquename=init]{biblatex}

\newcommand{\Description}[1]{}

\title{Zero-Shot Weight-Space Programming for Prediction and Classification of
Target Variables in Structured Datasets}
\author{Eduardo C. Garrido Merch\'an \\
Universidad Pontificia Comillas (IIT, DIGNUM) \\
Madrid, Spain \\
\texttt{ecgarrido@comillas.edu}}
\date{November 2026}

\begin{document}

\maketitle

\begin{abstract}
Weight-space programming fixes every parameter of a feed-forward network by hand, with no gradient descent anywhere in the construction. Every demonstration of the practice so far begins from a target function handed to the model as a rule system, which makes the construction a compilation problem. A supervised learning problem withholds that specification and exhibits its target only through a finite sample of labelled rows, so any procedure that produces weights for it must perform the statistical inference itself and then express the result in the parameterisation it was given. We ask whether a large language model can do both at once. The model is shown a small structured dataset in its entirety, in context, together with the architecture of a feed-forward network, and writes out every weight and bias as a literal number, with no optimizer, no gradient, no code execution, and no opportunity to evaluate the network before submitting it. Each construction is sealed under a digest and pre-registers its expected score before any evaluation runs. The suite is graded in four tiers: two toy problems whose optimal parameters fit on one line, nine synthetic problems ordered by standard data complexity measures and disguised by a hidden rotation, the classical tabular benchmarks, and one real engineering regression. The frozen network matches the closed-form optimum wherever that optimum is linear, reaches the irreducible error of the generating process on three rungs where it is not, and beats every fitted baseline on two of those, including the same architecture trained by gradient descent. It fails on two rungs, in both cases because the structure to be found is oblique to the coordinates it analysed. Three further constructors then separate two failure modes that a single score conflates: one returns no parameter vector at all on the hardest rungs while scoring well elsewhere, and another delivers everywhere while writing networks worse than predicting the mean and pre-registering that they will be good. One of the two diagnosed failures is shown to be avoidable, a second constructor solving the oblique-hinge rung by a margin that survives correction for multiple comparisons. What predicts whether the hidden layer earns its keep is the training error of a linear separator, which the composite complexity index fails to track.
\end{abstract}

\section{Introduction}

Gradient descent on data is how the field fixes the parameters of a neural
network. The representation itself is indifferent to their provenance. A single
hidden layer with a sigmoidal nonlinearity suffices to approximate any continuous
function on the unit cube to arbitrary accuracy
\citep{cybenko1989approximation}, and the rectified-linear networks used
throughout this paper represent arbitrary piecewise-linear functions once their
weights are fixed. Both statements concern which parameter vectors exist, and
both are silent about the procedure that finds one.
\citet{weightspaceprogrammingblog} exploited that silence directly, showing that
a large language model can read and write the parameters of a neural network as
literal numbers, with no optimizer anywhere in the loop. They named the practice
weight-space programming. In a companion study we carried it across a
difficulty-graded ladder of two-player games, constructing and independently
verifying ten hand-set networks that run from Nim to three-piece chess endgames
\citep{garridomerchan2026ladder}.

Every demonstration of the practice so far, ours included, begins from a target
function handed to the model as a rule system: the legal moves of a game, its
terminal conditions, and where one exists a closed-form optimal strategy. The
construction is therefore a compilation problem. The model is told what to
compute, and the difficulty lies in expressing a known computation through
linear maps and rectified units, which is why a small reusable vocabulary of
conjunctions, disjunctions, exact-match detectors and tiered scores carries an
entire ladder of games. A supervised learning problem withholds the
specification. Nobody can state which function of sepal length, sepal width,
petal length and petal width separates three species of iris, because no such
function is known in closed form; the function is exhibited only through a finite
sample of labelled rows. A procedure that produces weights for a target of this
kind must first perform the inference that identifies the function, and then
express the result in the parameterisation of the architecture it was given.
Those are two distinct abilities, and whether a language model has both at once,
with feedback on neither, has not been measured.

The question is worth answering because of what an externalised predictor makes
possible. The neighbouring literature on in-context learning shows that a model
shown labelled examples in its prompt behaves at inference time like a learner
over those examples \citep{brown2020language,garg2022what}, but the predictor it
infers remains inside the model, implicit in activations that exist only while
the prompt is being processed, and it must be re-elicited for every query. An
implicit predictor can be scored on the outputs it produces and on nothing else.
A parameter vector can be printed, audited coordinate by coordinate, deployed
without the language model, evaluated a billion times at the cost of a matrix
multiplication, and compared against the exact minimiser of the objective it is
supposed to approximate. That last comparison is the sharp one, and it is
unavailable to any study of an implicit predictor.

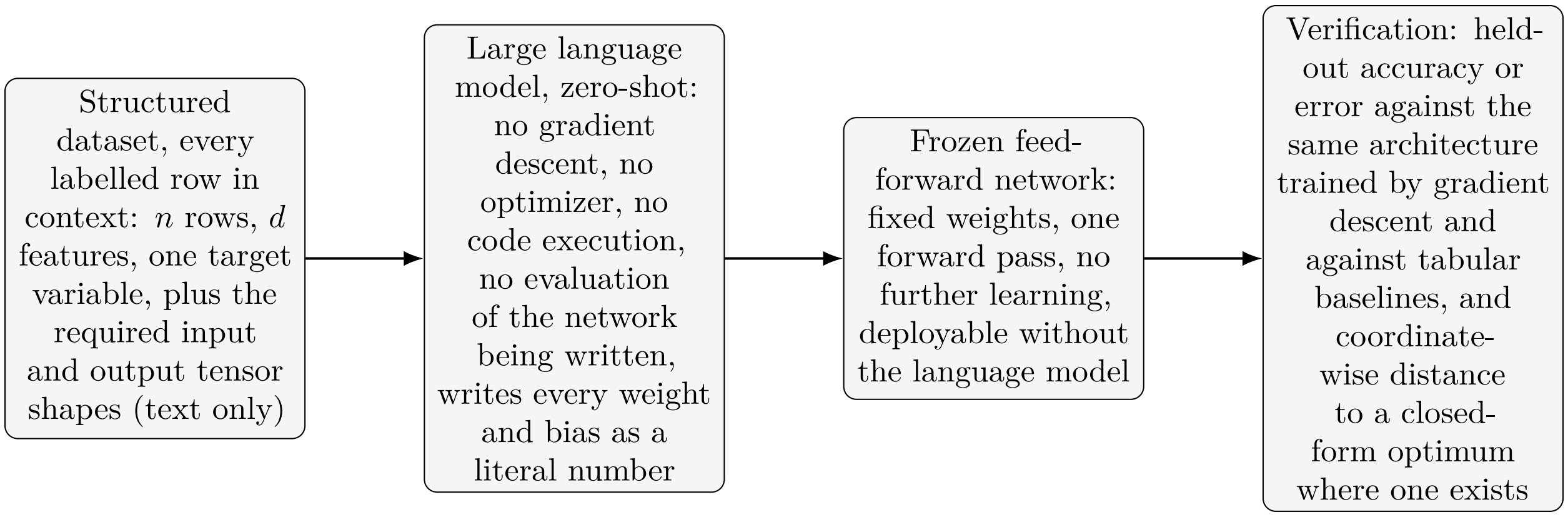
\begin{figure}[t]
\centering
\resizebox{\textwidth}{!}{%
\begin{tikzpicture}[
  node distance=0.9cm and 1.3cm,
  block/.style={draw, rounded corners, align=center, fill=gray!8, minimum width=3.3cm, minimum height=1.9cm, font=\small, text width=3.0cm},
  arr/.style={-{Latex[length=2.2mm]}, thick},
]
\node[block] (data) {Structured dataset, every labelled row in context: $n$ rows, $d$ features, one target variable, plus the required input and output tensor shapes (text only)};
\node[block, right=of data] (llm) {Large language model, zero-shot: no gradient descent, no optimizer, no code execution, no evaluation of the network being written, writes every weight and bias as a literal number};
\node[block, right=of llm] (net) {Frozen feed-forward network: fixed weights, one forward pass, no further learning, deployable without the language model};
\node[block, right=of net] (ver) {Verification: held-out accuracy or error against the same architecture trained by gradient descent and against tabular baselines, and coordinate-wise distance to a closed-form optimum where one exists};
\draw[arr] (data) -- (llm);
\draw[arr] (llm) -- (net);
\draw[arr] (net) -- (ver);
\end{tikzpicture}%
}
\caption{The capability test this paper applies under many conditions. A language model is shown a complete structured dataset and the required shapes of a feed-forward network, and must write out, from reasoning alone, every entry of that network's parameters so that its frozen forward pass predicts the target variable on held-out rows. Everything that varies across the paper, which dataset, whether that dataset is a classical one the model may have memorised or a synthetic one it cannot have seen, which architecture is being filled in, and which reference solution the result is compared against, is a controlled variation of this one test.}
\Description{A left-to-right pipeline of four boxes. The first box holds the structured dataset, every labelled row placed in the model's context together with the required input and output tensor shapes. An arrow leads to the second box, the language model operating zero-shot with no gradient descent, no optimizer, no code execution and no evaluation of the network it is writing, which emits every weight and bias as a literal number. An arrow leads to the third box, the frozen feed-forward network with fixed weights and a single forward pass. A final arrow leads to the fourth box, verification by held-out accuracy or error against the same architecture trained by gradient descent, against tabular baselines, and against a closed-form optimum where one exists.}
\label{fig:paradigm}
\end{figure}

We measure the capability on small structured datasets. The model is shown one
such dataset in its entirety, in context, together with the architecture of a
feed-forward network, and writes out every weight and bias as a literal number,
with no optimizer, no gradient, no code execution, and no opportunity to evaluate
the network before submitting it. Each construction is sealed under a
cryptographic digest and pre-registers, in words, the score it expects, before
any evaluation runs. Small tables are the right domain for three reasons. They
fit in context entirely, so the model is shown the data rather than a summary of
it. The comparison is informative in exactly this regime, because gradient-based
methods are at their weakest here and tree ensembles remain the state of the art
\citep{grinsztajn2022why,shwartzziv2022tabular}. And at a few hundred rows the
evaluation can be exhaustive rather than sampled. The suite itself is graded by a
measurement described in Section~\ref{sec:suite}: two toy problems whose optimal parameters fit on
one line, nine synthetic problems ordered by standard data complexity measures
and disguised by a hidden rotation, the classical tabular benchmarks, and one
real engineering regression. Figure~\ref{fig:paradigm} shows the test every
construction in the paper is produced and judged by.

The study contributes the following. First, a measurement of whether a language
model can perform statistical inference from a table and express the result
directly as the parameters of a feed-forward network, under a protocol that
forbids it from ever evaluating what it writes. Second, a difficulty-graded
suite of twelve rungs with sealed, pre-registered constructions, released with
the digests that fix the order of events. Third, a set of reference solutions
computed independently of the model, against which a written parameter vector is
compared coordinate by coordinate rather than only through the accuracy it
attains: the ordinary least-squares optimum where the objective is convex, the
Bayes-optimal rule and its irreducible error where we defined the generating
process ourselves, and Fisher's own 1936 discriminant coefficients on the Iris
measurements. Fourth, the identification of the two rungs where the practice
fails and of what those failures share, namely structure oblique to the
coordinates the constructor analysed. Fifth, three further constructors on the
identical suite, which separate two failure modes that a single score conflates,
and which show that one of the two diagnosed failures is avoidable. Sixth, the
finding that the quantity predicting whether a hidden layer earns its keep is the
training error of a linear separator, which the composite complexity index does
not track.

One threat to validity organises the design of the whole suite and has to be
named before any result is read. Iris, Wine and Breast Cancer Wisconsin appear
in textbooks, tutorials, package documentation and public repositories, and the
Iris measurements have been distributed as a named dataset since June 1988
\citep{fisher1936irisdataset}. \citet{bordt2024elephants} show that current
language models reproduce rows of exactly these datasets verbatim, and that
predictive performance is higher on datasets seen during training than on
datasets released afterwards. A score on an unmodified classical dataset
therefore cannot by itself distinguish inference from recall. The same work also
reports that models are surprisingly robust to transformations of memorised
data, so a rescaled or relabelled classical dataset is not a sufficient control
either. The synthetic tier, drawn from generating processes that did not exist
before this study, is accordingly the load-bearing evidence, and the classical
datasets are reported beside it rather than in place of it.

The paper is organized as follows. Section~\ref{sec:related} places the
construction against the four literatures it touches: in-context learning,
networks that emit the parameters of other networks, language models applied to
tabular prediction, and the data complexity measures that order the suite.
Section~\ref{sec:protocol} states the construction protocol, defines what counts
as weight-space programmed, and describes the sealing discipline that makes the
order of events checkable after the fact. Section~\ref{sec:suite} describes the
suite, its contamination controls, the complexity measurement that orders it, and
the baseline set. Section~\ref{sec:experiments} reports the measurements tier by
tier, then the pre-registered predictions against outcomes, then the three
further constructors and an open-weight size series.
Section~\ref{sec:discussion} draws the reading those results support, and
Section~\ref{sec:conclusion} closes with limitations and future work.

\section{Related work}
\label{sec:related}

The introduction sketched where this construction sits. This section gives the
positioning in detail, against four literatures the question touches, and the
construction sits at a different distance from each. We take the nearest first,
in-context learning, then networks that emit the parameters of other networks,
then language models applied to tabular prediction, and finally the data
complexity measures from which the suite takes its organizing axis.

The nearest is in-context learning. A large language model shown labelled examples in its prompt behaves at inference time like a learner over those examples \citep{brown2020language}, and the behaviour is sharp enough to admit a theory: transformers trained on synthetic linear regression problems learn in context to an accuracy approaching the best estimator for the function class they were trained on \citep{garg2022what}, and a single linear self-attention layer implements one step of gradient descent on a regression loss exactly, which exhibits a mechanism by which the behaviour can arise inside a trained transformer \citep{vonoswald2023transformers}. What that literature studies is a predictor that exists only while the prompt is being processed. The present setting asks for the same inference with the predictor externalised. An implicit predictor can be scored only on the outputs it produces, so the only available question is how accurate it is. An explicit one can be compared against the exact minimiser of the objective it is supposed to approximate, coordinate by coordinate, which is a different and sharper question and is the one this paper asks wherever a closed-form optimum exists.

The second is the literature on networks that emit the parameters of other networks. A hypernetwork produces the weights of a second network as the output of a first \citep{ha2017hypernetworks}, which is formally the same input-output relation as the one studied here. The generating network in that work is itself trained by gradient descent, and the pair is optimised end to end on the target task, so the weights it emits are the result of an optimisation over the very objective they are judged on. The model used here was never trained to emit weights, was never trained on the datasets in question, and receives no signal of any kind back from the network it produces. That nothing in the representational theory of feed-forward networks requires their parameters to arrive by gradient descent is a separate and older point: a single hidden layer with a sigmoidal nonlinearity approximates any continuous function on the unit cube to arbitrary accuracy \citep{cybenko1989approximation}, and the rectified-linear networks used throughout this paper represent arbitrary piecewise-linear functions once their parameters are fixed. Those are statements about which parameter vectors exist, and they are silent about the procedure that finds one.

The third is the application of language models to prediction on structured data, where the existing approaches keep the model inside the prediction loop. TabLLM serialises each row into a natural-language string and prompts the model for the label, obtaining non-trivial accuracy with no labelled examples at all and competing with gradient-boosted trees in the very-few-shot regime once the model is fine-tuned on a handful of rows \citep{hegselmann2023tabllm}. LIFT recasts non-language tasks, tabular classification and regression among them, as language modelling problems and fine-tunes on the resulting prompt-and-completion pairs \citep{dinh2022lift}. TabPFN avoids gradient descent on the target dataset by moving the training elsewhere, pretraining a transformer across millions of synthetic datasets drawn from a prior over tabular generating processes, after which it predicts on a new small dataset in a single forward pass; its authors report that it outperforms all previous methods on datasets of up to ten thousand samples, beating in under three seconds an ensemble of the strongest baselines given four hours of tuning \citep{hollmann2025accurate}. In all three the deployed predictor is a transformer and the dataset is present at inference time. The setting studied here removes the language model from deployment altogether. It is consulted once, emits a few hundred numbers, and is then discarded, after which the surviving artefact is a small network that could have been written by hand and that costs a matrix multiplication to evaluate.

The fourth is the literature this paper borrows its organising axis from. \citet{ho2002complexity} proposed measuring the difficulty of a classification problem by geometric descriptors of the labelled sample itself rather than by the error rate of any particular classifier, and introduced the families that have been used since: the overlap of the class-conditional ranges of individual features, the error of a linear separator and its behaviour on interpolated points, the fraction of points that lie on the boundary between classes in a minimum spanning tree, and the error of the nearest-neighbour rule. \citet{lorena2019complex} collected that family into a single normalised set, oriented so that every member lies in the unit interval and larger means harder, and \citet{lorena2018data} constructed the corresponding family for regression, where the analogues are the correlation of each feature with the output, the residual of a linear fit, and the smoothness of the output along a minimum spanning tree over the inputs. These measures are used here for the purpose they were designed for, which is to place problems on a common axis before any predictor is applied to them. They play no part in any construction, and they are computed after the constructions are sealed.

The threat to validity that organises the design of the suite is contamination. Iris, Wine and Breast Cancer Wisconsin appear in textbooks, tutorials, package documentation and public repositories, and the Iris measurements in particular have been in circulation since 1936 and have been distributed as a named dataset by a public machine learning repository since June 1988 \citep{fisher1936irisdataset}. \citet{bordt2024elephants} show that current language models reproduce rows of exactly these datasets verbatim, that few-shot predictive performance is higher on datasets seen during training than on datasets released afterwards, and that much of the performance observed on genuinely novel tabular problems is attributable to world knowledge rather than to statistical learning from the rows supplied. The same work reports that language models are surprisingly robust to transformations of memorised data, which is consistent with recall surviving a rescaling and which rules out treating a transformed classical dataset as a sufficient control. The synthetic tier is accordingly the load-bearing evidence of this paper and the classical datasets are reported beside it rather than in place of it. The same work also establishes the closest existing point of comparison for what follows: the sample efficiency of in-context few-shot prediction lags behind classical statistical learning algorithms, and the lag widens as the dimension of the problem grows. Whether the same lag appears when the model is asked to externalise its inference into weights, rather than to answer one query at a time, is the open question the present construction is designed to answer.

\section{The construction protocol}
\label{sec:protocol}

Section~\ref{sec:related} located the construction among its neighbours. This
section states what it is. We first fix the term, since the whole study turns on
a distinction from an ordinarily trained network, and then describe the four
steps of the protocol in the order they occur: the partition, the construction
itself, the sealing, and the architectures being filled in. The protocol fixes
the order of events and makes that order checkable afterwards, which matters more
here than in an ordinary empirical study, because the central claim is about what
the constructor did not see.

The definition is the companion study's, restated verbatim so that a reader
moving between the two papers meets a single formalism. It is stated
for an arbitrary task and makes no reference to games.

\begin{definition}
\label{def:wsp}
A feed-forward network $f_\theta$ is weight-space programmed for a task when
every entry of $\theta$ is fixed by a deterministic procedure, applied once
before $f_\theta$ is ever evaluated on an instance of the task, that takes as
input only a specification of the task and never takes as input a sample of the
network's own input-output behaviour. The procedure may be carried out by a
person, by a program, or by another neural network, and it may use arbitrary
non-differentiable computation, including exhaustive search over a finite space,
provided that the result of that computation enters $\theta$ as a constant and is
not invoked again at evaluation time.
\end{definition}

The definition transfers to the supervised setting without amendment, and the
reason is worth stating, because the data-dependence of the construction looks at
first like a violation of it. A labelled dataset is a specification of the task,
in the only form a supervised learning problem ever offers one. What the
procedure may never observe is what the network it is constructing does on any
input, and that prohibition is what rules out gradient descent, random search
over parameters, and the write-then-test-then-revise loop a model with code
execution would otherwise fall into. The line the definition draws separates
procedures that learn from the behaviour of the artefact they are building from
procedures that do not. Reading a table and writing down a decision boundary
falls on the permitted side of it; computing a gradient of the network's loss
does not.

The first step is the partition. A module that contains no construction and is executed before any construction exists draws every dataset, fixes a stratified partition under a single seed, and serialises only the training half into a prompt file. For the twelve rungs introduced in this paper the seed is the date the suite was built and the split is in half, except on the real problem where three hundred rows go into the prompt and the remaining seven hundred and five are held out. The prompt carries the task description, the encoding of the network's inputs and outputs, the two architectures to be filled in, and every training row. It carries no summary statistic of any kind. The test suite re-derives the prompt from the partition and compares it to the file on disk byte for byte, and separately checks that no held-out feature vector appears anywhere in it.

The second step is the construction. The model reads one prompt file and writes one module containing every weight and every bias as a literal number, together with a docstring recording the derivation and a prediction, in words, of what the resulting network will score. Nothing else is permitted. The model executes no code on the data, computes no gradient, fits nothing, and above all never evaluates the network it is writing: the arithmetic is done by reading the table. The prohibition on evaluation is the one that does the work, because a model with the ability to score its own network and revise it would be performing a search over parameters guided by the network's behaviour, which is the thing the definition excludes. What the model is permitted to do, and does throughout, is check its own rule against rows it was given, which the definition permits, since it reads the table and never the network.

The construction modules are constrained syntactically so that this cannot be circumvented. A module may contain only literal numbers, a metadata dictionary and a docstring. It may not import anything, may not call any function, may not contain a loop or a comprehension, may not access an attribute, and may not reference any name other than its own four parameter arrays. The test suite enforces this by walking the abstract syntax tree of every module, so a construction that attempted to read the dataset at import time would fail to load rather than quietly succeed.

The third step is sealing. A script records the SHA-256 digest of every construction file, its size and its line count, and refuses to reseal a file whose digest has changed. Only after the seal is written does any evaluation run. Every number reported in this paper is reproduced by the test suite directly from the sealed file, and the digest of that file is printed beside it in the released results. A construction revised after its score was known would therefore have to be revised before the seal, which is the event the digest records.

Two architectures are filled in on every rung. The first, which we call \texttt{linear}, is a single affine map from the raw features to the output, with no hidden layer: a weight matrix and a bias vector, and nothing else. The second, \texttt{mlp8}, inserts one hidden layer of eight rectified units between the same input and the same output. Both are evaluated in double precision with every parameter marked non-trainable, and both the hand-written and the gradient-trained arm of every comparison instantiate the same class from the same module, so no difference between them can come from a difference in the forward pass. Classification rungs emit one logit per class and predict by argmax; regression rungs emit a single real number and are scored by squared error. The features are presented in raw units with no standardisation anywhere, which is a deliberate difficulty: a construction that ignores the scale of a column whose values run into the hundreds cannot work, and the gradient-trained arm is reported both on raw inputs, which is the encoding the constructor was given, and on standardised inputs, which is what a practitioner would do and which is the stronger of the two baselines on most rungs.

The constructing model is Claude Opus 5 for every construction reported here, except those of the final experimental subsection, which puts three further constructors through the identical protocol on the identical suite. Every construction is one pass per rung and architecture, with no iteration and no ensembling, at the reasoning effort the harness records as \texttt{xhigh}, one step above its \texttt{high} setting and one below its highest. That setting is taken from the session log, which records it per request and independently of anything the constructor says about itself. The constructions disagree with it and with each other: the sealed metadata of the twelve rungs introduced here describes the setting as "high" and the metadata of an earlier round describes it as "ultrahigh", neither of which is a setting the harness offers. Both are self-reports, the log is a record, and a sealed module may not be edited once its digest is written, so the discrepancy is stated here and the modules are left exactly as they were sealed. The numbers in this paper are therefore not the best of several attempts, and they carry no claim about any other model or any other effort setting.

\section{The suite, and how its rungs are ordered}
\label{sec:suite}

Section~\ref{sec:protocol} fixed how a construction is produced and sealed. What
remains to fix is what it is asked to construct for, and a suite assembled by
taste would make any finding about where the practice stops working
uninterpretable. This section describes the four tiers of the suite, the
contamination controls applied to each, the measurement that orders the synthetic
tier on an axis independent of our judgement, and the baselines every rung is
scored against. The suite holds four tiers and twenty-four hand-written
constructions across the twelve rungs introduced here, together with the sixteen
constructions of the benchmark tier carried over from the earlier study.

The toy tier has two rungs. In \texttt{toy\_line} a single feature determines a single target through an affine relation with a small amount of noise, over forty rows; the optimal parameter vector has two entries and the ordinary least-squares solution is available in closed form. In \texttt{toy\_gap} two features determine one of two groups, over forty-eight rows, with the two groups separated by eight within-group standard deviations so that the Bayes error is fifteen parts in a million. These two rungs exist so that the mechanics of the construction can be displayed in full, and so that the apparatus has a case in which anything other than a near-perfect score would indicate a fault in the apparatus rather than a limitation of the method.

The synthetic tier has nine rungs, five of classification and four of regression, each drawn from a generating process defined for this study and each accompanied by a sealed record of that process. On the classification side, \texttt{s1\_sep} places two groups in four dimensions with a hard margin along an unknown direction, so the Bayes error is exactly zero and the only question is whether a hand-written rule can find the direction. \texttt{s2\_lin} replaces the margin with an overlap, drawing two Gaussians with a shared non-diagonal covariance whose separation is solved for rather than guessed, so that the Bayes error is exactly ten per cent. \texttt{s3\_xor} places four clusters at the corners of a square with the labels assigned by parity, which makes the two class-conditional means coincide and renders every rule built from a difference of means useless. \texttt{s4\_ball} surrounds a compact core with a shell, so the optimal boundary is closed and no half-plane can enclose it. \texttt{s5\_spiral} winds two arms a turn and a quarter around a common centre, which is the first boundary in the suite whose curvature changes faster than a small number of half-planes can follow. On the regression side, \texttt{r1\_lin} is exactly linear in five features; \texttt{r2\_kink} is a sum of three rectified hinges, which places the optimum exactly inside the \texttt{mlp8} hypothesis class and makes the gap to it pure estimation error; \texttt{r3\_inter} is a bilinear saddle, which no rectified network represents exactly; and \texttt{r4\_sine} is two full periods of a sinusoid in a single feature, where eight rectified units buy eight breakpoints and the question is where to spend them.

Every synthetic rung is disguised before it is shown. The sample is drawn in latent coordinates, then rotated by a Haar-distributed orthogonal matrix, then scaled per column by a factor drawn log-uniformly over two orders of magnitude, then shifted, and for the classification rungs the class labels are permuted. The rotation is the operative part. The same model that wrote the constructions also authored the generating module, so it knows that \texttt{s3\_xor} is a parity problem and that \texttt{s4\_ball} is radial; knowing that a two-dimensional problem is a checkerboard does not tell anyone which two directions the checkerboard is aligned with, and a hand-written weight matrix has to name those directions numerically. No numerical parameter of any rung, no rotation, no scale, no label permutation and no irreducible error rate is visible anywhere the constructor reads, and the test suite checks that no value from the sealed record appears in the corresponding prompt. The descriptions shown above each table are deliberately uninformative, disclosing the number of rows, the number of columns and nothing else.

The benchmark tier is the suite of the earlier study, reported here unchanged: the Iris measurements, an anonymised image of the same one hundred and fifty plants under a hidden affine map with hidden column and label permutations, a six-dimensional three-class synthetic problem with correlated features and a known Bayes rule, the Wine chemistry measurements whose columns span four orders of magnitude, and Breast Cancer Wisconsin, whose training half contains eight thousand five hundred and twenty numbers.

The real tier is a single rung, \texttt{concrete}, holding one thousand and five distinct laboratory mixtures of concrete, eight ingredient and curing-time columns, and the compressive strength each mixture reached in megapascals. No part of this study generated these rows, no closed-form optimum exists for them, and the relation between the ingredients and the strength is known to be nonlinear in curing age. Two properties of the raw table are dealt with in the partition rather than discovered later. Twenty-five records duplicate another record in all nine columns and are dropped; a further thirteen mixtures appear more than once with the same ingredient quantities and a different measured strength, which are genuine replicate experiments and are kept, with the partition taken over distinct mixtures rather than over rows so that no ingredient vector appears on both sides of it.

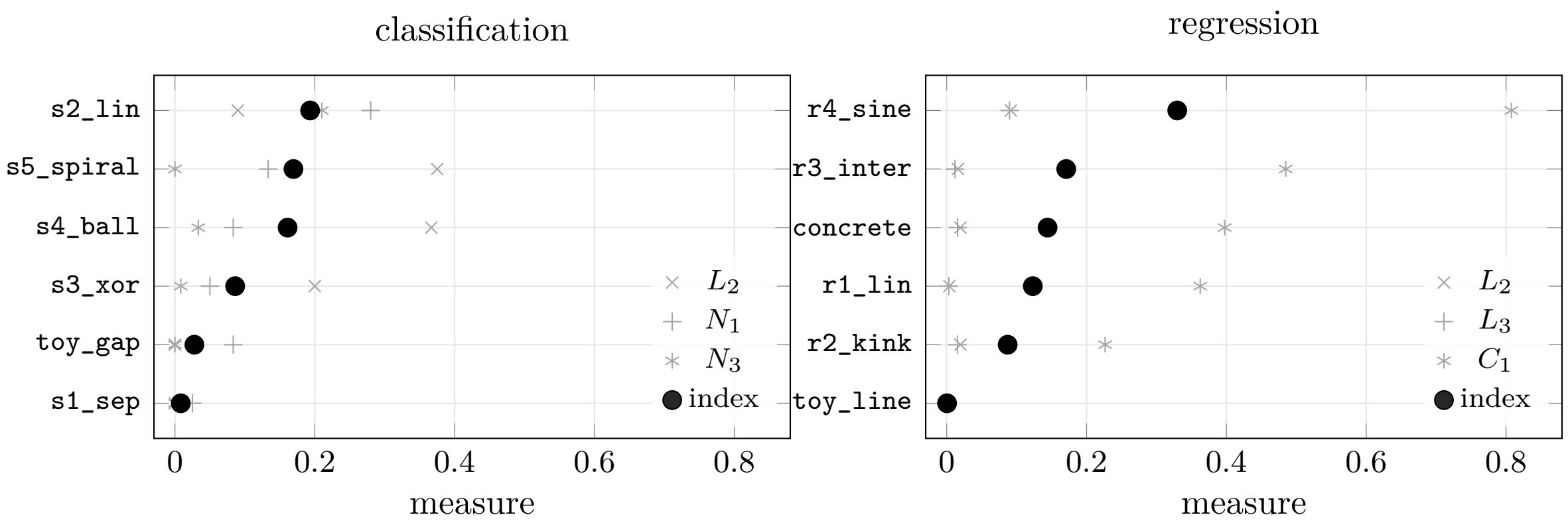
\begin{figure}[t]
\centering
\begin{tikzpicture}
\pgfplotsset{
  ladderstyle/.style={width=0.48\textwidth, height=5.2cm, xmin=-0.03, xmax=0.88,
    tick label style={font=\footnotesize}, label style={font=\small},
    title style={font=\small}, grid=both, grid style={gray!18},
    ytick={0,...,5}, ymin=-0.6, ymax=5.6,
    yticklabel style={font=\scriptsize}},
}
\begin{axis}[ladderstyle, name=left, xlabel={measure}, title={classification},
  yticklabels={\texttt{s1\_sep},\texttt{toy\_gap},\texttt{s3\_xor},\texttt{s4\_ball},
    \texttt{s5\_spiral},\texttt{s2\_lin}},
  legend style={font=\scriptsize, at={(0.98,0.03)}, anchor=south east, draw=none,
    fill=white, fill opacity=0.85, text opacity=1}]
\addplot[only marks, mark=x, mark size=2.4pt, gray!70] table[x=m1,y=i] {figdata/difficulty_clf.dat};
\addlegendentry{$L_2$}
\addplot[only marks, mark=+, mark size=2.6pt, gray!70] table[x=m2,y=i] {figdata/difficulty_clf.dat};
\addlegendentry{$N_1$}
\addplot[only marks, mark=asterisk, mark size=2.2pt, gray!70] table[x=m3,y=i] {figdata/difficulty_clf.dat};
\addlegendentry{$N_3$}
\addplot[only marks, mark=*, mark size=2.6pt, black] table[x=index,y=i] {figdata/difficulty_clf.dat};
\addlegendentry{index}
\end{axis}
\begin{axis}[ladderstyle, name=right, at={(left.east)}, xshift=1.35cm, anchor=west,
  xlabel={measure}, title={regression},
  yticklabels={\texttt{toy\_line},\texttt{r2\_kink},\texttt{r1\_lin},\texttt{concrete},
    \texttt{r3\_inter},\texttt{r4\_sine}},
  legend style={font=\scriptsize, at={(0.98,0.03)}, anchor=south east, draw=none,
    fill=white, fill opacity=0.85, text opacity=1}]
\addplot[only marks, mark=x, mark size=2.4pt, gray!70] table[x=m1,y=i] {figdata/difficulty_reg.dat};
\addlegendentry{$L_2$}
\addplot[only marks, mark=+, mark size=2.6pt, gray!70] table[x=m2,y=i] {figdata/difficulty_reg.dat};
\addlegendentry{$L_3$}
\addplot[only marks, mark=asterisk, mark size=2.2pt, gray!70] table[x=m3,y=i] {figdata/difficulty_reg.dat};
\addlegendentry{$C_1$}
\addplot[only marks, mark=*, mark size=2.6pt, black] table[x=index,y=i] {figdata/difficulty_reg.dat};
\addlegendentry{index}
\end{axis}
\end{tikzpicture}
\caption{The suite ordered by measured difficulty, with the easiest rung at the bottom of each panel. The filled marker is the composite complexity index, the mean of the three open markers beside it, and every measure is normalised so that larger is harder and is computed from the training rows alone after the constructions on that rung were sealed. The two task types are drawn separately because their headline triples differ: for classification the training error of a linear separator, the fraction of points on a class boundary in a minimum spanning tree, and the leave-one-out error of the nearest-neighbour rule; for regression the mean squared residual of the least-squares fit, that residual on interpolated points, and the correlation deficit of the best single feature. The disagreement between a mean and its members is visible directly. On \texttt{s3\_xor} the linear separator fails on a fifth of the rows while the nearest-neighbour rule fails on under one per cent, which places the rung low on the index and high on the only measure that predicts what a hidden layer is worth; on \texttt{r4\_sine} all three members agree, and that rung is the hardest in the suite by every one of them.}
\Description{Two horizontal dot plots side by side, classification on the left and regression on the right, with six rungs each ordered from easiest at the bottom to hardest at the top. Each rung carries four markers: a filled one for the composite complexity index and three open ones for the measures it averages. On most rungs the three open markers lie close together, but on the parity rung one of them is far to the right of the other two, and on the hardest regression rung all three agree.}
\label{fig:difficulty}
\end{figure}

The ordering of the synthetic tier is a design claim, and we check it against a measurement that does not depend on us. For each rung we compute, from the training rows alone and only after every construction on that rung is sealed, the standard complexity measures for its task type: for classification the five feature-overlap measures, the three linearity measures, the six neighbourhood measures, the three dimensionality measures and the two class-balance measures collected by \citet{lorena2019complex}; for regression the four correlation measures, the three linearity measures, the four smoothness measures and the dimensionality measure of \citet{lorena2018data}. Three of them carry the ordering in the tables that follow, chosen because they bear directly on what a small feed-forward network can express. For classification these are the training error of a linear separator, the fraction of points incident to an edge of the minimum spanning tree that joins two classes, and the leave-one-out error of the nearest-neighbour rule; for regression they are the mean squared residual of the least-squares fit, the same residual evaluated on points interpolated between output-adjacent examples, and the correlation deficit of the most informative single feature. The mean of the three is reported as a single complexity index, and the implementation is tested against fixtures whose answers are known in advance, including the one that distinguishes a parity problem from a noise problem: a checkerboard has almost no nearest-neighbour error and almost no boundary points while remaining far from linearly separable, and a sample with random labels has all three. Figure~\ref{fig:difficulty} shows the suite ordered by that index, which exhibits the grading of the ladder, and it already makes one point the experiments return to: the index is a mean, its members disagree, and the rung it ranks easiest among the five classification problems with any difficulty at all is the parity rung, whose linear separator error is the second largest in the tier.

\section{Experiments}
\label{sec:experiments}

The suite of Section~\ref{sec:suite} is now run. The six subsections that follow
take the four tiers in order of difficulty, from the toy problems through the
synthetic tier, the classical benchmarks and the real regression, then compare
each construction's pre-registered prediction against its outcome, and close by
varying the constructor: three further models on the identical suite, and a
size-graded series of open-weight models that locates the floor of the
capability. Every number below is the score of a frozen parameter vector that was written out by hand, sealed under a digest, and only then evaluated. The baselines are the same architectures trained by Adam over twenty seeds, ordinary least squares or penalised logistic regression where the objective is convex, linear and quadratic discriminant analysis, a nearest-neighbour rule, a random forest and gradient-boosted trees, all fitted on the same training rows and scored on the same held-out rows. Classification is scored by accuracy with Wilson intervals and paired exact McNemar tests against the median seed of each baseline; regression is scored by the coefficient of determination with a paired bootstrap over rows on the difference in squared error. Where the generating process is known, the irreducible error and, for the two-dimensional classification rungs, the accuracy of the best half-plane obtained by exhaustive search over directions and thresholds are reported beside the measured scores.

\subsection{Toy problems}

The toy tier exists to display the mechanics and to give the apparatus a case in which anything short of a near-perfect score would indicate a fault in the apparatus. Both rungs were solved exactly.

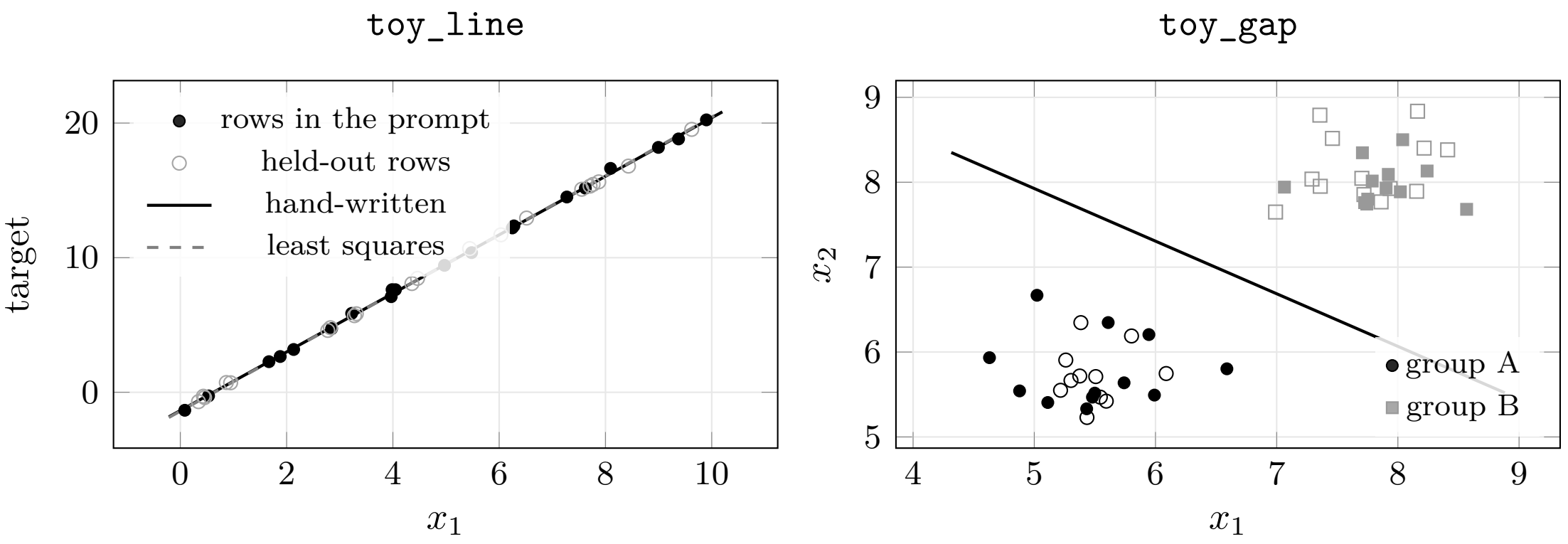
\begin{figure}[t]
\centering
\begin{tikzpicture}
\begin{axis}[
  name=left, width=0.47\textwidth, height=5.0cm,
  xlabel={$x_1$}, ylabel={target}, tick label style={font=\footnotesize},
  label style={font=\small}, legend style={font=\scriptsize, at={(0.03,0.97)},
  anchor=north west, draw=none, fill=white, fill opacity=0.85,
  text opacity=1}, grid=both, grid style={gray!18},
  title={\small \texttt{toy\_line}}, title style={font=\small},
]
\addplot[only marks, mark=*, mark size=1.5pt, black] table[x=x,y=y] {figdata/toy_line_train.dat};
\addlegendentry{rows in the prompt}
\addplot[only marks, mark=o, mark size=1.8pt, gray!70] table[x=x,y=y] {figdata/toy_line_test.dat};
\addlegendentry{held-out rows}
\addplot[thick, black] table[x=x,y=hand] {figdata/toy_line_fits.dat};
\addlegendentry{hand-written}
\addplot[thick, dashed, gray] table[x=x,y=ols] {figdata/toy_line_fits.dat};
\addlegendentry{least squares}
\end{axis}
\begin{axis}[
  name=right, at={(left.east)}, xshift=1.1cm, anchor=west,
  width=0.47\textwidth, height=5.0cm,
  xlabel={$x_1$}, ylabel={$x_2$}, tick label style={font=\footnotesize},
  label style={font=\small}, legend style={font=\scriptsize, at={(0.97,0.03)},
  anchor=south east, draw=none, fill=white, fill opacity=0.85,
  text opacity=1}, grid=both, grid style={gray!18},
  title={\small \texttt{toy\_gap}}, title style={font=\small},
]
\addplot[only marks, mark=*, mark size=1.5pt, black] table[x=x,y=y] {figdata/toy_gap_train_A.dat};
\addlegendentry{group A}
\addplot[only marks, mark=square*, mark size=1.5pt, gray!80] table[x=x,y=y] {figdata/toy_gap_train_B.dat};
\addlegendentry{group B}
\addplot[only marks, mark=o, mark size=1.8pt, black] table[x=x,y=y] {figdata/toy_gap_test_A.dat};
\addplot[only marks, mark=square, mark size=1.8pt, gray!80] table[x=x,y=y] {figdata/toy_gap_test_B.dat};
\addplot[thick, black, forget plot] table[x=x,y=y] {figdata/toy_gap_boundary.dat};
\end{axis}
\end{tikzpicture}
\caption{The two toy rungs with the hand-written solutions drawn over them. Filled markers are the rows placed in the prompt and hollow markers the held-out rows, which the constructing model never saw. On the left the hand-written line and the least-squares line computed from the same twenty rows are separated by less than the width of the stroke. On the right the hand-written boundary separates the forty-eight specimens without error; the closest held-out row lies more than two within-group standard deviations from it.}
\Description{Two scatter plots side by side. The left plot shows a one-dimensional regression problem with twenty training points and twenty held-out points lying almost exactly on a straight line, with the hand-written fit and the least-squares fit indistinguishable. The right plot shows two well-separated clusters in two dimensions with a straight decision boundary passing through the empty gap between them.}
\label{fig:toy}
\end{figure}

On \texttt{toy\_line} the model read twenty rows of one feature and one target, split them at the median of the feature, summed both columns within each half by hand, and took the slope through the two group means. Figure~\ref{fig:toy} shows the result. The written slope is $2.175233$ and the written intercept is $-1.356180$, against an ordinary least-squares solution of $2.184284$ and $-1.400799$ computed from the same twenty rows, a coordinate-wise distance of $0.0091$ and $0.0446$. The generating process, sealed before the construction existed, used a slope of $2.152457$ and an intercept of $-1.208310$, so the hand-written vector is in fact nearer the truth than the least-squares estimate is, which is a fact about the noise in this particular sample and carries no implication for the estimator. On the held-out rows the hand-written network reaches a coefficient of determination of $0.9997$ against $0.9997$ for least squares, with root mean squared errors of $0.1046$ and $0.1166$ against a noise standard deviation of $0.150$.

On \texttt{toy\_gap} the model computed the two class means and the pooled within-class variances by hand, read the within-class correlation off the table and found it negligible, and wrote the diagonal Gaussian discriminant with its boundary at the midpoint of the two means. It predicted twenty-four of twenty-four correct before any held-out row was read, and scored twenty-four of twenty-four.

\begin{figure}[t]
\centering
\begin{tikzpicture}[
  box/.style={draw, rounded corners, align=left, fill=gray!6, font=\footnotesize,
              inner sep=5pt, text width=7.1cm, minimum height=2.5cm},
  arr/.style={-{Latex[length=2.2mm]}, thick},
  node distance=0.55cm and 0.55cm,
]
\node[box] (a) {{\scshape What the prompt shows}\\[3pt]
  Twenty rows of one feature and one target, in raw units. No summary statistic
  of any kind, and no held-out row.};
\node[box, right=of a] (b) {{\scshape What the model computes}\\[3pt]
  Sort on $x_1$, split at the median, sum both columns in each half:\\[3pt]
  \hspace*{1em}lower ten: $\sum x_1 = 24.3656$, $\sum y = 39.4391$\\
  \hspace*{1em}upper ten: $\sum x_1 = 74.2416$, $\sum y = 147.9310$};
\node[box, below=of a] (c) {{\scshape What the model derives}\\[3pt]
  $\text{slope} = (14.79310 - 3.94391)/(7.42416 - 2.43656) = 2.175233$\\[3pt]
  $\text{intercept} = 9.368505 - 2.175233 \times 4.930360 = -1.356180$};
\node[box, right=of c] (d) {{\scshape What the model writes}\\[3pt]
  \hspace*{1em}\texttt{W1 = [[2.175233]]}\\
  \hspace*{1em}\texttt{b1 = [-1.356180]}\\[3pt]
  and, for \texttt{mlp8}, the same map carried through one rectified unit held
  twenty units from its kink.};
\draw[arr] (a) -- (b);
\draw[arr] (b.south) -- ++(0,-0.27) -| (c.north);
\draw[arr] (c) -- (d);
\end{tikzpicture}
\caption{The complete \texttt{toy\_line} construction, which is the smallest instance in the paper of what every other rung does at scale. Nothing between the first box and the last was executed: the four sums were taken by reading the table, the two divisions were done in the model's reasoning, and the two numbers that came out were written into a module that contains no import, no call and no name other than its own parameter arrays. The same module declares, before the seal, that it expects a coefficient of determination between $0.995$ and $0.9995$; it scored $0.9997$.}
\Description{A left-to-right diagram of four boxes. The first describes the prompt, which holds twenty raw rows and no statistics. The second shows the four column sums the model computed by hand after splitting the rows at the median. The third shows the slope and intercept derived from those sums by two divisions. The fourth shows the two literal numbers written into the construction module.}
\label{fig:weights}
\end{figure}
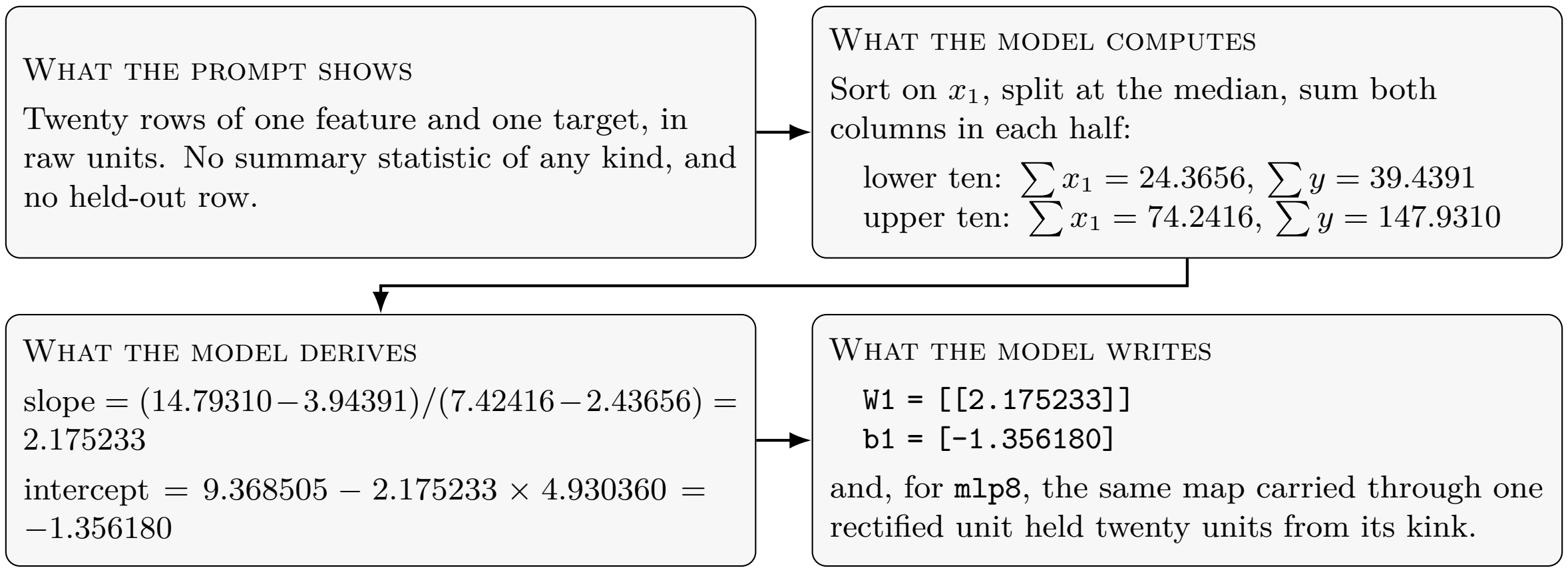

Figure~\ref{fig:weights} reproduces the whole of the \texttt{toy\_line} construction, from the sums the model wrote down to the parameters that came out of them, which is the smallest complete instance of what the rest of the paper does at scale.

\subsection{Synthetic problems of increasing complexity}

The nine synthetic rungs are the load-bearing evidence of the paper, because their rows did not exist before this study and their optimal parameters are known only to the sealed record. Table~\ref{tab:synth} reports them. Four findings separate out, and they do not point the same way.

The first is that where the optimal rule lies inside the architecture and the construction can find it, the hand-written network reaches the irreducible error. On \texttt{s1\_sep} the model identified the separating direction, discarded the one column whose contrast was within a standard error of zero, applied a two-column correlation correction that nearly doubled the weight on the second informative feature, and classified eighty of eighty held-out rows without error, which is the Bayes rate of a rung constructed to have none. On \texttt{s2\_lin} it solved a three-by-three system by hand from covariances read off the table and scored $0.8800$ against an irreducible rate of $0.9000$; penalised logistic regression scored $0.9400$ on the same rows, which is above the Bayes rate and is sampling luck in a test half of a hundred.

The second is that the hidden layer is used when, and only when, the table shows a reason to use it. On five of the nine rungs the model declined the hidden layer outright and wrote an exact re-encoding of its own linear rule, carrying the affine map through one rectified unit held far from its kink and leaving the other seven provably inert. The test suite confirms that the two architectures then compute the same function to within $10^{-13}$ on every row of every one of those five datasets, and the measured scores are identical to four decimal places. Where the table did show a reason, the units were spent and the gain was large: $24.2$ points of accuracy on the parity rung, $37.5$ on the radial rung, $0.624$ of coefficient of determination on the saddle and $1.046$ on the sinusoid.

The third is that two of those constructions reached the ceiling and one exceeded every baseline. On \texttt{s3\_xor} the model recovered the four cluster centres and verified that they form a parallelogram, by checking that its diagonals share a midpoint. It derived the two boundary directions as the sides of that parallelogram. It then computed the parity rule exactly rather than approximately, using the identity $(p+q)^2 - (p-q)^2 = 4pq$ to turn a product into a difference of two absolute values, and each absolute value into two rectified units. The result classifies all one hundred and twenty held-out rows correctly, above the Bayes rate of $0.9877$ estimated on four hundred thousand fresh rows, and above every baseline including the trained network of identical architecture, which reached $0.9662$. On \texttt{s4\_ball} the model fenced the core with an octagon of eight rectified units and observed that summing the hinges rather than taking their maximum rounds the corners, so that the boundary stays within one per cent of a circle in every direction; the test suite verifies this by bisecting along a thousand rays. It scored $0.9917$ against an irreducible $0.9896$. On \texttt{r4\_sine} the model located the period from the four zero crossings and estimated the amplitude by averaging the target over the rows bracketing each extremum, rather than by taking the largest value in the table. It then showed that the obvious triangle-wave fit wastes eight kinks, derived the optimal trapezoid instead, and placed its knots half a radian either side of each extremum. It scored $0.9460$ against an irreducible ceiling of $0.9500$, beating the best baseline at $0.9312$.

The fourth is that two constructions failed, and in both cases the cause is nameable and is a property of the procedure rather than of the effort. On \texttt{r2\_kink} the optimum lies exactly inside the architecture, which is the most favourable situation in the whole suite, and the construction nonetheless reached $0.5084$ against $0.8594$ for the trained network of the same shape and $0.9123$ for the generating network itself. The reason is visible once the sealed record is opened. The model located the kinks by banding on one feature and regressing on the other, which can only find kinks whose boundaries are parallel to a coordinate axis, and the three true hinge directions are $(0.049, -0.237)$, $(0.542, 0.175)$ and $(0.336, -0.216)$, none of them axis-aligned. A procedure that reads one feature at a time cannot see an oblique fold, and this is the rung on which that limitation is fatal. On \texttt{s5\_spiral} the construction scored $0.5750$, which is below its own linear arm at $0.6167$ and is the only case anywhere in this study where spending the hidden layer made the network worse. The model built a triangle wave in a tilted coordinate $p = x_1 - 0.10 x_2$, with stripe boundaries fitted from two bands of the table. Opening the record shows that the tilt is right where the arms are far apart and wrong where they are not: in the band of $x_2$ furthest from the spiral's centre the construction classifies fourteen of fourteen rows correctly, and in the band containing the centre it classifies seven of thirty-four, which is worse than chance because the inner stripes come out inverted. The failure is a direct consequence of representing a curve of varying pitch by a fixed set of parallel half-planes, which the construction itself predicted would be its weakness and then underestimated.

\begin{figure}[t]
\centering
\begin{tikzpicture}
\begin{axis}[
  width=0.52\textwidth, height=5.4cm,
  xlabel={$L_2$ of the training rows}, ylabel={gain from the hidden layer},
  tick label style={font=\footnotesize}, label style={font=\small},
  grid=both, grid style={gray!18}, legend style={font=\scriptsize,
  at={(0.97,0.97)}, anchor=north east, draw=none, fill=white,
  fill opacity=0.85, text opacity=1},
  xmin=-0.02, xmax=0.42,
]
\addplot[only marks, mark=*, mark size=2.2pt, black] table[x=l2, y expr={\thisrow{mlp}-\thisrow{lin}}] {figdata/ladder_clf.dat};
\addlegendentry{classification (accuracy)}
\addplot[only marks, mark=square*, mark size=2.2pt, gray!75] table[x=l2, y expr={\thisrow{mlp}-\thisrow{lin}}] {figdata/ladder_reg.dat};
\addlegendentry{regression ($R^2$)}
\addplot[domain=0:0.42, thin, dashed, black] {x};
\addlegendentry{$y = L_2$}
\end{axis}
\end{tikzpicture}
\caption{What the hidden layer buys, against how far from linearly separable the training rows are. For the classification rungs the two quantities agree closely over the range in which the construction succeeds: the hand-written linear arm loses almost exactly the fraction of points that no half-plane can place correctly, and the hidden layer recovers almost exactly that fraction. The one classification point below the line is \texttt{s5\_spiral}, where the hidden layer was spent and lost accuracy. The regression points sit far above the line because the residual of a least-squares fit on a target scaled to the unit interval is not on the same scale as a difference of coefficients of determination; only their ordering is comparable.}
\Description{A scatter plot with the linear-separator training error on the horizontal axis and the gain in score from adding a hidden layer on the vertical axis. Classification points lie close to the diagonal line y equals x for four of the five rungs, with one point well below the line. Regression points lie above the line and increase with the horizontal coordinate.}
\label{fig:gap}
\end{figure}
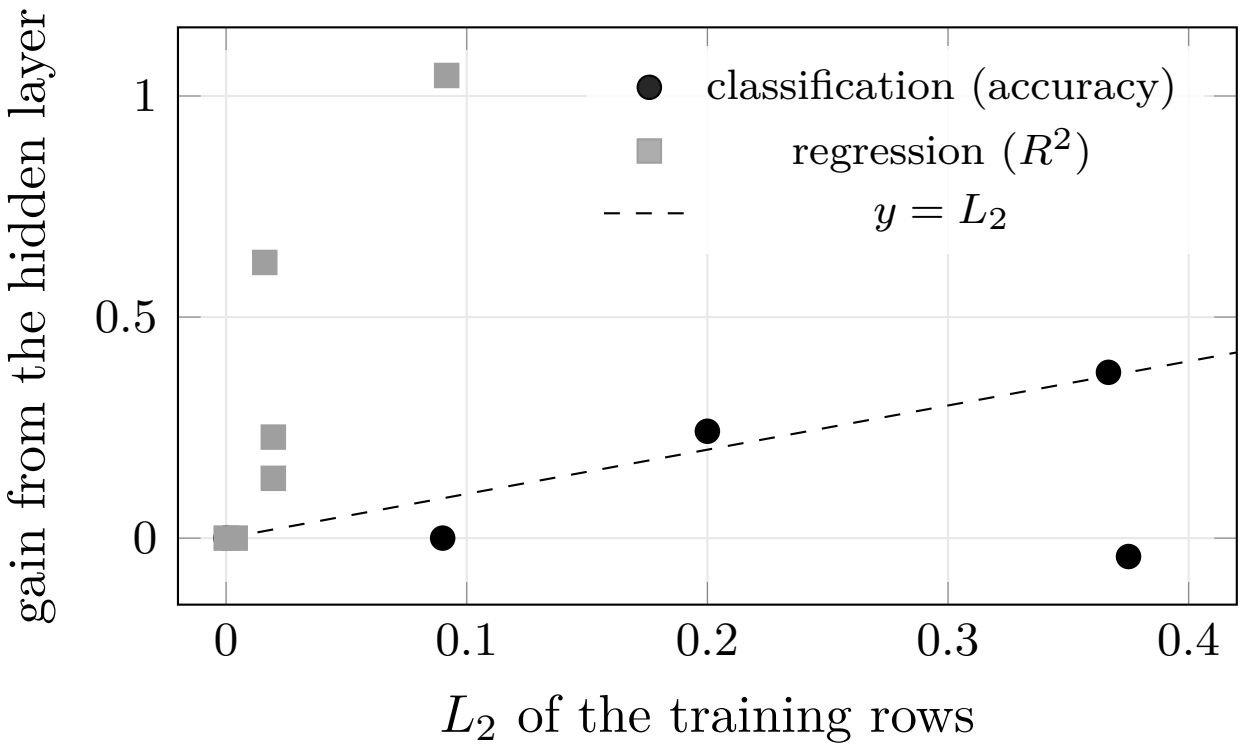

Figure~\ref{fig:regions} draws the three nonlinear classification boundaries, and Figure~\ref{fig:gap} relates the value of the hidden layer to the measured complexity of each rung.

\begin{table}[t]
\caption{The nine synthetic rungs. ``Linear'' and ``MLP'' are the hand-written constructions for the two architectures; ``best baseline'' is the strongest of the twelve fitted predictors on that rung; ``ceiling'' is one minus the irreducible error for classification and the attainable coefficient of determination for regression, both computed from the sealed generating process; ``half-plane'' is the accuracy of the best half-plane found by exhaustive search, reported for the two-dimensional rungs where the optimal boundary is not linear. ``$L_2$'' is the training error of a linear separator for classification and the mean squared residual of the least-squares fit for regression, computed from the training rows only and after every construction was sealed. Bold marks the three rungs on which a hand-written construction matches or beats every one of the twelve fitted baselines.}
\label{tab:synth}
\small
\begin{tabular}{llrrrrrrr}
\toprule
rung & task & $n$ & $d$ & $L_2$ & linear & MLP & best baseline & ceiling \\
\midrule
\texttt{s1\_sep}    & classification & 80/80   & 4 & 0.000 & \textbf{1.0000} & \textbf{1.0000} & 1.0000 & 1.0000 \\
\texttt{s2\_lin}    & classification & 100/100 & 4 & 0.090 & 0.8800 & 0.8800 & 0.9400 & 0.9000 \\
\texttt{s3\_xor}    & classification & 120/120 & 2 & 0.200 & 0.7583 & \textbf{1.0000} & 0.9917 & 0.9877 \\
\texttt{s4\_ball}   & classification & 120/120 & 2 & 0.367 & 0.6167 & 0.9917 & 1.0000 & 0.9896 \\
\texttt{s5\_spiral} & classification & 120/120 & 2 & 0.375 & 0.6167 & 0.5750 & 0.9917 & 0.9997 \\
\midrule
\texttt{r1\_lin}    & regression & 100/100 & 5 & 0.004 & 0.8872 & 0.8872 & 0.9117 & 0.9004 \\
\texttt{r2\_kink}   & regression & 100/100 & 2 & 0.020 & 0.3731 & 0.5084 & 0.8594 & 0.9180 \\
\texttt{r3\_inter}  & regression & 120/120 & 2 & 0.016 & 0.2813 & 0.9048 & 0.9068 & 0.9192 \\
\texttt{r4\_sine}   & regression & 80/80   & 1 & 0.092 & $-0.1004$ & \textbf{0.9460} & 0.9312 & 0.9500 \\
\bottomrule
\end{tabular}
\\[4pt]
\small The half-plane ceilings, where they apply, are $0.7250$ for \texttt{s3\_xor}, $0.6526$ for \texttt{s4\_ball} and $0.6076$ for \texttt{s5\_spiral}; the three hand-written linear arms scored $0.7583$, $0.6167$ and $0.6167$ against them.
\end{table}

\begin{figure}[t]
\centering
\begin{tikzpicture}
\pgfplotsset{
  regionstyle/.style={width=0.355\textwidth, height=4.5cm, axis equal image=false,
    tick label style={font=\tiny}, label style={font=\footnotesize},
    title style={font=\small}, xtick=\empty, ytick=\empty},
}
\begin{axis}[regionstyle, name=p1, title={\texttt{s3\_xor}: $1.0000$}]
\addplot[only marks, mark=*, mark size=1.0pt, black] table[x=x,y=y] {figdata/s3_xor_A.dat};
\addplot[only marks, mark=square*, mark size=1.0pt, gray!70] table[x=x,y=y] {figdata/s3_xor_B.dat};
\addplot[thick, black] table[x=x,y=y] {figdata/s3_xor_boundary.dat};
\end{axis}
\begin{axis}[regionstyle, name=p2, at={(p1.east)}, xshift=0.7cm, anchor=west,
  title={\texttt{s4\_ball}: $0.9917$}]
\addplot[only marks, mark=*, mark size=1.0pt, black] table[x=x,y=y] {figdata/s4_ball_A.dat};
\addplot[only marks, mark=square*, mark size=1.0pt, gray!70] table[x=x,y=y] {figdata/s4_ball_B.dat};
\addplot[thick, black] table[x=x,y=y] {figdata/s4_ball_boundary.dat};
\end{axis}
\begin{axis}[regionstyle, name=p3, at={(p2.east)}, xshift=0.7cm, anchor=west,
  title={\texttt{s5\_spiral}: $0.5750$}]
\addplot[only marks, mark=*, mark size=1.0pt, black] table[x=x,y=y] {figdata/s5_spiral_A.dat};
\addplot[only marks, mark=square*, mark size=1.0pt, gray!70] table[x=x,y=y] {figdata/s5_spiral_B.dat};
\addplot[thick, black] table[x=x,y=y] {figdata/s5_spiral_boundary.dat};
\end{axis}
\end{tikzpicture}
\caption{Decision boundaries of the three hand-written \texttt{mlp8} constructions whose rungs are not linearly separable, drawn over all two hundred and forty rows of each, with the held-out accuracy in the title. The parity boundary is the exact pair of lines through the centre of the cluster parallelogram; the radial boundary is an octagon whose corners the sum of its own hinges has rounded to within one per cent of a circle; the spiral boundary is a set of parallel stripes in a tilted coordinate, which tracks the arms where they are far apart and inverts where they wind in.}
\Description{Three scatter plots. The first shows four clusters at the corners of a tilted parallelogram with two straight lines crossing at its centre, separating the clusters by parity. The second shows a compact core inside a surrounding ring with a nearly circular closed boundary between them. The third shows two interleaved spiral arms crossed by a set of parallel straight stripes that follow the arms near the outside of the figure and cut across them near the centre.}
\label{fig:regions}
\end{figure}
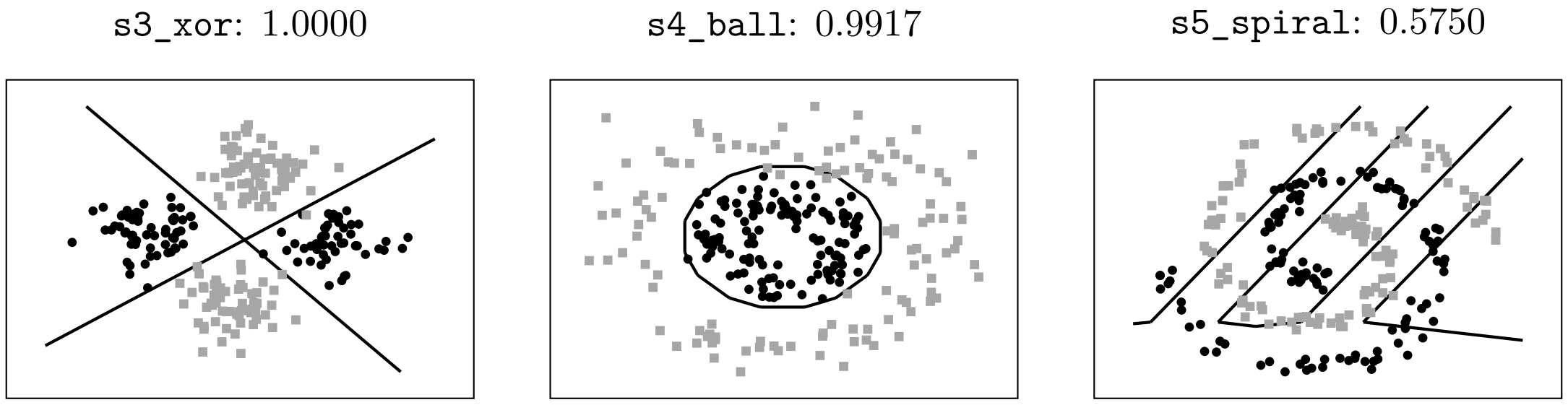

\subsection{Benchmark datasets}

The benchmark tier is the suite of the companion study, reported here for comparability and unchanged. Under the same no-code-execution condition the hand-written linear constructions score $0.9200$ on Iris, $0.9600$ on its anonymised counterpart, $0.7133$ on the six-dimensional correlated synthetic problem, $0.9663$ on Wine and $0.9509$ on Breast Cancer Wisconsin; the eight-unit constructions score $0.9200$, $0.9600$, $0.7133$, $0.9775$ and $0.9439$. Averaged over the five datasets, the hand-written network is $1.18$ accuracy points below the same architecture trained on raw inputs for the linear arm, and $0.02$ points below it for the eight-unit arm. It sits between zero and seven tenths of a point below gradient-boosted trees, random forests and the nearest-neighbour rule, and between two and three and a half points below the convex optima of logistic regression and linear discriminant analysis. A paired exact McNemar test finds no significant difference against any baseline on Iris, on its anonymised counterpart or on Wine; the significant differences in that tier are concentrated on the correlated synthetic problem, where the construction assumed a diagonal covariance and said so in writing before it was scored.

Two features of that tier bear on the present one. The first is the memorisation control. The hand-written network scores four points higher on the anonymised Iris table than on the recognisable one, which is the opposite of the ordering recall would produce. The construction for the anonymised table records in writing that the model did suspect the provenance of the rows, and explains why the suspicion is useless once the scale, the column order and the label assignment are all hidden. The second is that the correlated synthetic rung of that tier, where the construction fell $7.3$ points short of the Bayes rule, failed for the same reason \texttt{r2\_kink} fails here: the model declined to estimate and invert a covariance matrix by eye, and the structure it therefore could not see was precisely the one that mattered.

\subsection{A real problem}

The real tier is a single regression rung on one thousand and five distinct concrete mixtures, three hundred of which were placed in the prompt. It is the largest table in the study by a wide margin, at two thousand seven hundred numbers, and the only one with no generating process and no closed-form optimum.

The construction is the most elaborate in the paper and the most clearly physical. The model identified curing age as the dominant effect and read its shape from matched pairs of rows that differ only in age. The gains it found are $20.6$ megapascals from one day to seven, $11.3$ from seven days to twenty-eight, $7.1$ from twenty-eight to ninety and $0.7$ from ninety to a hundred and eighty. It fitted these as a piecewise-linear curve with knots at three, seven, twenty-eight and ninety days. It identified three further nonlinearities and spent one unit on each: a penalty for mixtures whose binder content, counting slag at four fifths of the rate of cement, falls below two hundred and eighty kilograms; a floor on the benefit of reducing water below a hundred and sixty kilograms, justified by the two driest mixtures in the table being eight and fifteen megapascals weaker than a straight extrapolation predicts; and a saturation of the cement effect above four hundred kilograms. The linear part of the mixture and the base slope of the age curve ride together on a single always-active rectifier, which is what makes eight units enough. The construction scores $0.7444$ against $0.7892$ for the same architecture trained by gradient descent, $0.8749$ for gradient-boosted trees, $0.8096$ for a random forest and $0.5697$ for ordinary least squares. The linear arm, obtained by projecting the same additive model onto a plane rather than by refitting, scores $0.5159$ against $0.5697$ for the least-squares optimum it is approximating.

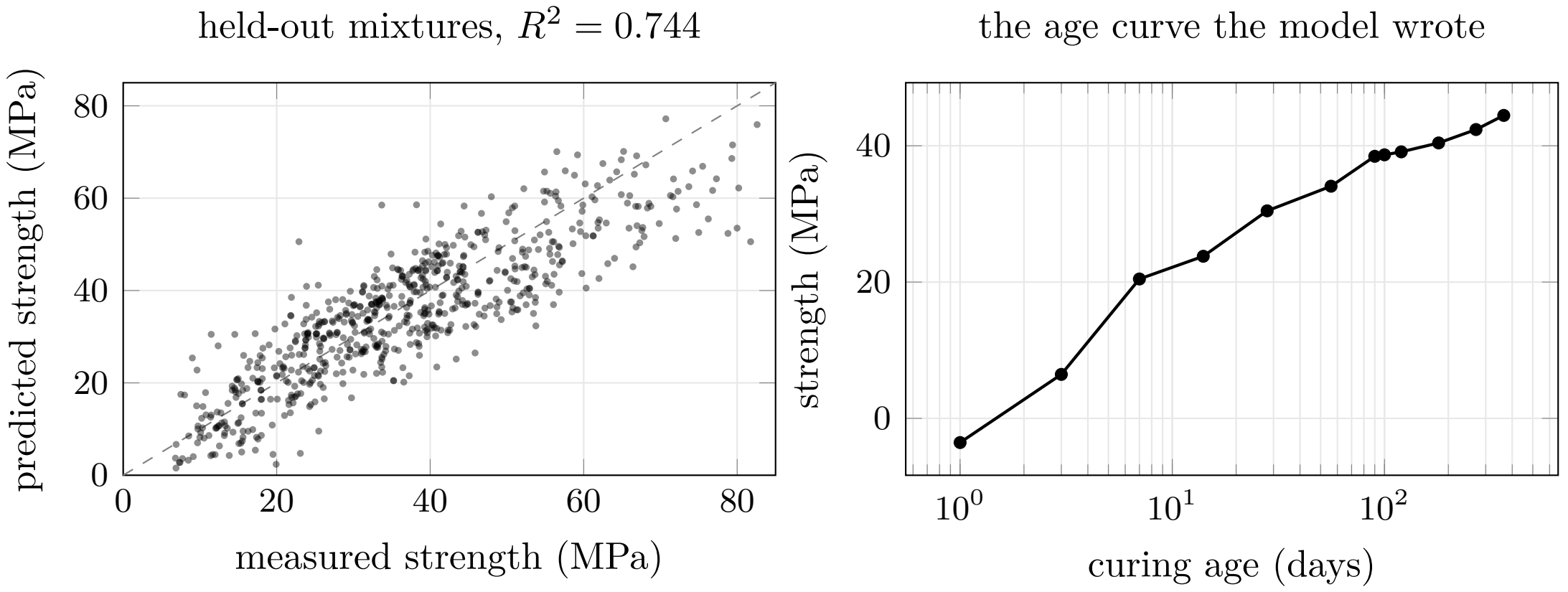
\begin{figure}[t]
\centering
\begin{tikzpicture}
\begin{axis}[
  name=left, width=0.46\textwidth, height=5.2cm,
  xlabel={measured strength (MPa)}, ylabel={predicted strength (MPa)},
  tick label style={font=\footnotesize}, label style={font=\small},
  grid=both, grid style={gray!18}, xmin=0, xmax=85, ymin=0, ymax=85,
  title={\small held-out mixtures, $R^2=0.744$}, title style={font=\small},
]
\addplot[only marks, mark=*, mark size=0.7pt, black, opacity=0.45] table[x=obs,y=pred] {figdata/concrete_pred.dat};
\addplot[domain=0:85, thin, dashed, gray] {x};
\end{axis}
\begin{axis}[
  name=right, at={(left.east)}, xshift=1.2cm, anchor=west,
  width=0.46\textwidth, height=5.2cm,
  xlabel={curing age (days)}, ylabel={strength (MPa)}, xmode=log,
  tick label style={font=\footnotesize}, label style={font=\small},
  grid=both, grid style={gray!18},
  title={\small the age curve the model wrote}, title style={font=\small},
]
\addplot[thick, black, mark=*, mark size=1.3pt] table[x=age,y=strength] {figdata/concrete_age_curve.dat};
\end{axis}
\end{tikzpicture}
\caption{The real rung. On the left, the hand-written eight-unit network's predictions against the measured compressive strength of the seven hundred and five held-out mixtures, with the identity drawn for reference; the spread is wider at the top of the range, which is where the multiplicative age effect the construction could not represent does the most damage. On the right, the piecewise-linear age curve the model derived from matched pairs of rows, evaluated at the median mixture and drawn on a logarithmic age axis, on which a logarithmic law would be a straight line.}
\Description{Two panels. The left panel is a scatter plot of predicted against measured concrete strength for seven hundred and five held-out mixtures, clustered around the identity line with wider spread at high strengths. The right panel plots the fitted age curve against a logarithmic age axis from one day to a year, rising steeply at first and flattening after about ninety days.}
\label{fig:real}
\end{figure}

The gap to gradient boosting is real and the construction named its cause before it was measured: the effect of curing age is multiplicative in the strength of the mixture and additive in this model, so a rich dry mixture at three days is over-predicted by seven to fifteen megapascals, and representing a product of two learned functions is exactly what a hidden layer of rectified units cannot do while a tree ensemble does it for free. Figure~\ref{fig:real} shows the predictions against the measurements and the age curve the construction wrote.

\subsection{Pre-registered predictions against outcomes}

Every construction states, in its own docstring and before the seal, the score it expects. Five of the twenty-four declare no interval at all but an identity, namely that they reproduce their own linear counterpart exactly, and all five hold: the two networks agree to within $10^{-13}$ on every row of every one of those five datasets and their measured scores are equal. Of the nineteen that declare a numeric interval, nine landed inside it. Of the ten that did not, six were pessimistic and four optimistic, which continues the pattern the companion study reported: the model underestimates what it will achieve more often than it overestimates it. The pessimistic misses are concentrated where the construction is exact and the model treated its own exactness as approximate, predicting between $0.92$ and $0.98$ for a parity rule that is an identity and scoring $1.0000$, and predicting between $0.65$ and $0.85$ for the saddle and scoring $0.9048$. The optimistic misses are concentrated on the two failures already described, where the construction predicted $0.72$ to $0.85$ on the spiral and scored $0.5750$, and $0.85$ to $0.93$ on the hinge rung and scored $0.5084$. The three largest misses in the study are therefore all cases in which the model's self-assessment and its actual limitation came apart, and in two of the three the model had already written down the correct mechanism and then mis-sized its effect.

\subsection{Four constructors on the same suite}

Everything that surrounds the construction is fixed: the partition, the prompts, the two architectures, the twelve fitted baselines, the sealing discipline and the scoring code. That is what makes the suite usable for the question the constructions themselves cannot answer, which is what changes when the model writing the weights changes. We report three further arms here, written by Claude Sonnet 5, by Claude Haiku 4.5 and by Grok 4.7, each under the identical brief, with one fresh constructor per rung carrying no memory of any other rung, no code execution, and no access to the sealed records, to any other arm's modules or to any result file. Every submission passed the same syntax-tree gate, was sealed under its own digest before any score existed, and was scored by the same harness on the same held-out rows. An exploratory fifth attempt, an eight-billion-parameter Qwen 3 served locally, preceded the pre-registered round and is reported here only to say why it was discarded: it processed every prompt in full and stopped of its own accord, and the modules it wrote declare parameters of the wrong shape, a model identity of \enquote{human} and a prediction of \enquote{$15.0 \pm 5.0$} for a coefficient of determination, so no module from it was ever admitted, sealed or scored. That attempt carried no stopping rule and no order of administration, which is what Section~\ref{sec:openweight} supplies: a graded series of five open-weight models under a rule fixed before the first dispatch, which locates the floor of this capability rather than observing one model fall below it.

Three asymmetries govern how the numbers below may be read, and all three are released per rung in a dispatch log alongside the modules. The first is that the Sonnet arm covers nine of the twelve rungs, for reasons taken up below, while the other two cover all twelve. The second is that three of those nine Sonnet rungs were produced under an instruction the other six never carried, namely an explicit budget on the length of the reply, imposed after the constructor had twice exhausted the harness's output ceiling and returned nothing at all; the Haiku and Grok arms carried no such instruction on any rung and needed no second dispatch anywhere. The third asymmetry is the most consequential and runs against the result rather than for it. The arms were not equally isolated. The Opus arm reported in the preceding subsections was written by the same session that had authored the generating processes, and it was trusted not to consult them. The Sonnet and Haiku arms were written by subagents whose only tool was a file reader pointed at one kit file. The Grok arm was written through a command line with its tool allow-list set to the empty string, so that constructor had no file access of any kind and could not have reached the sealed records had it tried. Isolation therefore increases in the order Opus, then Sonnet and Haiku, then Grok, and any ordering of results that followed that same sequence would be uninterpretable. Table~\ref{tab:arms} shows that the results do not follow it, because the arm with the strictest isolation is the one that scores highest.

\begin{table}[t]
\caption{The four constructor arms on all twelve rungs, each entry the better of that arm's two architectures on the same held-out rows. ``Ceiling'' is one minus the irreducible error for classification and the attainable coefficient of determination for regression, both from the sealed generating process; the real rung has neither. ``Base'' is the strongest of the twelve fitted predictors. A dash marks a rung for which that constructor returned nothing across four dispatches at two output ceilings. The means at the foot are taken over the nine rungs every arm delivered, which is the only set on which the four are comparable with one another. Bold marks the highest entry of each row among the four arms, and every arm that ties it.}
\label{tab:arms}
\small
\begin{tabular}{lrrrrrrr}
\toprule
rung & $L_2$ & ceiling & base & Opus 5 & Grok 4.7 & Sonnet 5 & Haiku 4.5 \\
\midrule
\texttt{toy\_line}   & 0.000 & 0.9994 & 0.9997 & \textbf{0.9997} & \textbf{0.9997} & \textbf{0.9997} & 0.9996 \\
\texttt{toy\_gap}    & 0.000 & 1.0000 & 1.0000 & \textbf{1.0000} & \textbf{1.0000} & \textbf{1.0000} & \textbf{1.0000} \\
\texttt{s1\_sep}     & 0.000 & 1.0000 & 1.0000 & \textbf{1.0000} & \textbf{1.0000} & 0.9750 & 0.9625 \\
\texttt{s2\_lin}     & 0.090 & 0.9000 & 0.9400 & 0.8800 & \textbf{0.9100} & 0.8800 & 0.7800 \\
\texttt{s3\_xor}     & 0.200 & 0.9877 & 0.9917 & \textbf{1.0000} & 0.9333 & 0.6333 & 0.6500 \\
\texttt{s4\_ball}    & 0.367 & 0.9896 & 1.0000 & 0.9917 & 0.9833 & \textbf{1.0000} & 0.5750 \\
\texttt{s5\_spiral}  & 0.375 & 0.9997 & 0.9917 & 0.6167 & \textbf{0.7417} & 0.6833 & 0.6167 \\
\texttt{r1\_lin}     & 0.004 & 0.9004 & 0.9117 & \textbf{0.8872} & 0.8858 & 0.8631 & 0.1718 \\
\texttt{r2\_kink}    & 0.020 & 0.9180 & 0.8594 & 0.5084 & \textbf{0.8448} & --- & 0.2511 \\
\texttt{r3\_inter}   & 0.016 & 0.9192 & 0.9068 & \textbf{0.9048} & 0.8451 & --- & $-0.6593$ \\
\texttt{r4\_sine}    & 0.092 & 0.9500 & 0.9312 & \textbf{0.9460} & 0.9289 & 0.0336 & 0.0338 \\
\texttt{concrete}    & 0.020 & --- & 0.8749 & \textbf{0.7444} & 0.7410 & --- & 0.3554 \\
\midrule
delivered & & & & 12 & 12 & 9 & 12 \\
mean over the nine common & & & & 0.9246 & \textbf{0.9314} & 0.7853 & 0.6433 \\
\bottomrule
\end{tabular}
\end{table}

The first result is that delivery and quality are separate failures, and the two weaker arms fail in different ways. Sonnet returned nothing at all on \texttt{r2\_kink}, \texttt{r3\_inter} and \texttt{concrete}, each dispatched four times to fresh constructors at two different output ceilings. The two dispatches whose token profile was recorded are identical across all three rungs and across both ceilings: four consecutive completions truncated at the ceiling, every one consumed inside the reasoning block, with no module text emitted at any point, for half a million output tokens per rung at the higher of the two settings. Doubling the ceiling doubled what was spent and moved the constructor no closer to delivering, which rules out the length of the reply as the binding constraint and leaves non-convergence on those rungs under this protocol. Haiku, by contrast, delivered everywhere and was willing to write a network whose predictions are worse than the mean of the training targets, reaching $-0.6593$ on the saddle and $-0.3575$ on the sinusoid. A practitioner choosing between these two behaviours is choosing between a constructor that declines to answer and one that answers badly without saying so, and the second is the more dangerous because its output passes every structural check this protocol applies.

The second result is that the two strongest arms agree on when the hidden layer is worth spending, to the rung. Where the table gives no reason to bend the decision boundary, a constructor can decline the hidden layer honestly by carrying its affine map through one rectified unit held far from its kink and leaving the other seven provably inert. Opus and Grok each do this on exactly the same five rungs, \texttt{toy\_line}, \texttt{toy\_gap}, \texttt{s1\_sep}, \texttt{s2\_lin} and \texttt{r1\_lin}, and on no others; the test suite confirms in both cases that the two architectures then compute the same function to within $10^{-13}$ on every held-out row. Sonnet declines on those five and on the sinusoid as well, which is the one rung in the suite where declining is the error. Haiku declines on seven, and its set is wrong in both directions at once: it spends the hidden layer on \texttt{r1\_lin}, whose optimum is exactly linear and where the spending costs it $0.23$ of coefficient of determination, and it declines on \texttt{s4\_ball} and \texttt{s5\_spiral}, whose optima are a closed curve and a pair of spiral arms. Knowing when a hidden layer is useless is therefore a competence that separates these constructors, and the two arms that have it are the two that score highest.

The third result is what survives a correction for multiple comparisons. Each pair of arms is tested on up to twenty-four rung-and-architecture combinations at once, by an exact McNemar test on the discordant rows for classification and a paired bootstrap on the difference in squared error for regression, and at the five per cent level more than one of those would be expected to separate by chance alone. We therefore apply the Holm step-down correction within each family and report both the uncorrected and the corrected counts. Against Sonnet, Opus is ahead on five comparisons uncorrected and on three after correction, and behind on none. Against Haiku, Opus is ahead on twelve uncorrected and on nine after correction, and behind on none. Against Grok the picture is a tie that the correction sharpens rather than removes: four comparisons each way uncorrected, and after correction one in Opus's favour and two in Grok's, with twenty-one of the twenty-four not separated at all. Table~\ref{tab:holm} lists every comparison that survives.

\begin{table}[t]
\caption{Every paired comparison that survives the Holm correction within its family, with the first arm's score minus the second's on the same held-out rows. Classification is compared by an exact McNemar test on the discordant rows and regression by a paired bootstrap on the difference in squared error. Comparisons not listed were not separated after correction, including four that reach $p<0.05$ uncorrected and do not survive, of which the largest is the spiral rung discussed in the text.}
\label{tab:holm}
\small
\begin{tabular}{llrrl}
\toprule
family & rung and architecture & $\Delta$ & $p$ & in favour of \\
\midrule
Opus against Grok    & \texttt{r2\_kink}, MLP    & $-0.3364$ & $<0.0001$ & Grok \\
                     & \texttt{r3\_inter}, linear & $-0.1537$ & $0.0006$  & Grok \\
                     & \texttt{r3\_inter}, MLP    & $+0.0598$ & $0.0010$  & Opus \\
\midrule
Opus against Sonnet  & \texttt{s3\_xor}, linear   & $+0.2000$ & $<0.0001$ & Opus \\
                     & \texttt{s3\_xor}, MLP      & $+0.3667$ & $<0.0001$ & Opus \\
                     & \texttt{r4\_sine}, MLP     & $+0.9125$ & $<0.0001$ & Opus \\
\midrule
Opus against Haiku   & \texttt{s3\_xor}, MLP      & $+0.4417$ & $<0.0001$ & Opus \\
                     & \texttt{s4\_ball}, MLP     & $+0.4167$ & $<0.0001$ & Opus \\
                     & \texttt{r1\_lin}, linear   & $+0.7155$ & $<0.0001$ & Opus \\
                     & \texttt{r1\_lin}, MLP      & $+0.9471$ & $<0.0001$ & Opus \\
                     & \texttt{r3\_inter}, linear & $+0.9982$ & $<0.0001$ & Opus \\
                     & \texttt{r3\_inter}, MLP    & $+1.5642$ & $<0.0001$ & Opus \\
                     & \texttt{r4\_sine}, MLP     & $+1.3035$ & $<0.0001$ & Opus \\
                     & \texttt{concrete}, linear  & $+0.1606$ & $<0.0001$ & Opus \\
                     & \texttt{concrete}, MLP     & $+0.3890$ & $<0.0001$ & Opus \\
\bottomrule
\end{tabular}
\end{table}

The fourth result revises a claim made earlier in this paper. The preceding subsections diagnosed two failures of the Opus arm, on the hinge rung and on the spiral, as failures of frame: the structure to be found is oblique to the coordinates the construction analysed, and no name for it was available within the vocabulary of a fixed number of parallel half-planes. A second constructor shown the same rows contradicts the second half of that diagnosis on one of the two rungs and points the same way on the other. On \texttt{r2\_kink} Grok reaches $0.8448$ against Opus's $0.5084$, a difference that survives the Holm correction with a bootstrap interval on the difference in mean squared error of $[0.592, 1.365]$, so the oblique folds that defeated one construction were found by another and the limitation is a property of the procedure rather than of the architecture or of the rung. On \texttt{s5\_spiral} Grok reaches $0.7417$ against $0.5750$, and Opus is the only arm whose hidden layer costs it accuracy relative to its own linear rule there, but that difference does not survive correction and we therefore record it as suggestive and unestablished. The honest summary is that one of the two diagnosed failures is now known to be avoidable and the other is probably avoidable.

The fifth result concerns what the constructions say about themselves before they are scored. Every module pre-registers, in its own docstring and before the seal, an interval for the score it expects, and those intervals are audited against the measured scores in a released file. Grok lands inside its interval on twenty of twenty-four modules, Sonnet on sixteen of eighteen, Opus on nine of the nineteen that state an interval rather than an exact identity with its own linear counterpart, and Haiku on four of twenty-four. The direction of the misses matters more than the count. Opus misses pessimistically more often than optimistically, and Grok's four misses are all smaller than five hundredths. Haiku's twenty misses are optimistic seventeen times, and the overstatements are large rather than marginal: it predicted a coefficient of determination between $0.76$ and $0.82$ on the saddle and scored $-0.6593$, and between $0.55$ and $0.68$ on the sinusoid and scored $-0.3575$. Calibration and capability therefore do not simply track one another across these four arms, since Opus is the least calibrated of the three strong arms while tying for the best score, but the weakest arm is the only one that is systematically and largely overconfident, and it is overconfident precisely where its networks are worse than predicting the mean.

Two further observations do not fit the pattern and are recorded because they do not. Sonnet achieves $1.0000$ on the radial rung, the best score any arm reached there, above the irreducible $0.9896$ of the generating process and level with the strongest fitted baseline, on a single discordant row against Opus; the difference is not separated and the result is reported for what it is, which is a reminder that an arm that fails to deliver on three rungs is not uniformly weaker everywhere else. And on the overlapping-Gaussian rung Opus and Sonnet predict identically on all one hundred held-out rows in both architectures, having estimated covariances by hand from the same table, which is an agreement of two independent constructions at the level of the prediction rather than of the score.

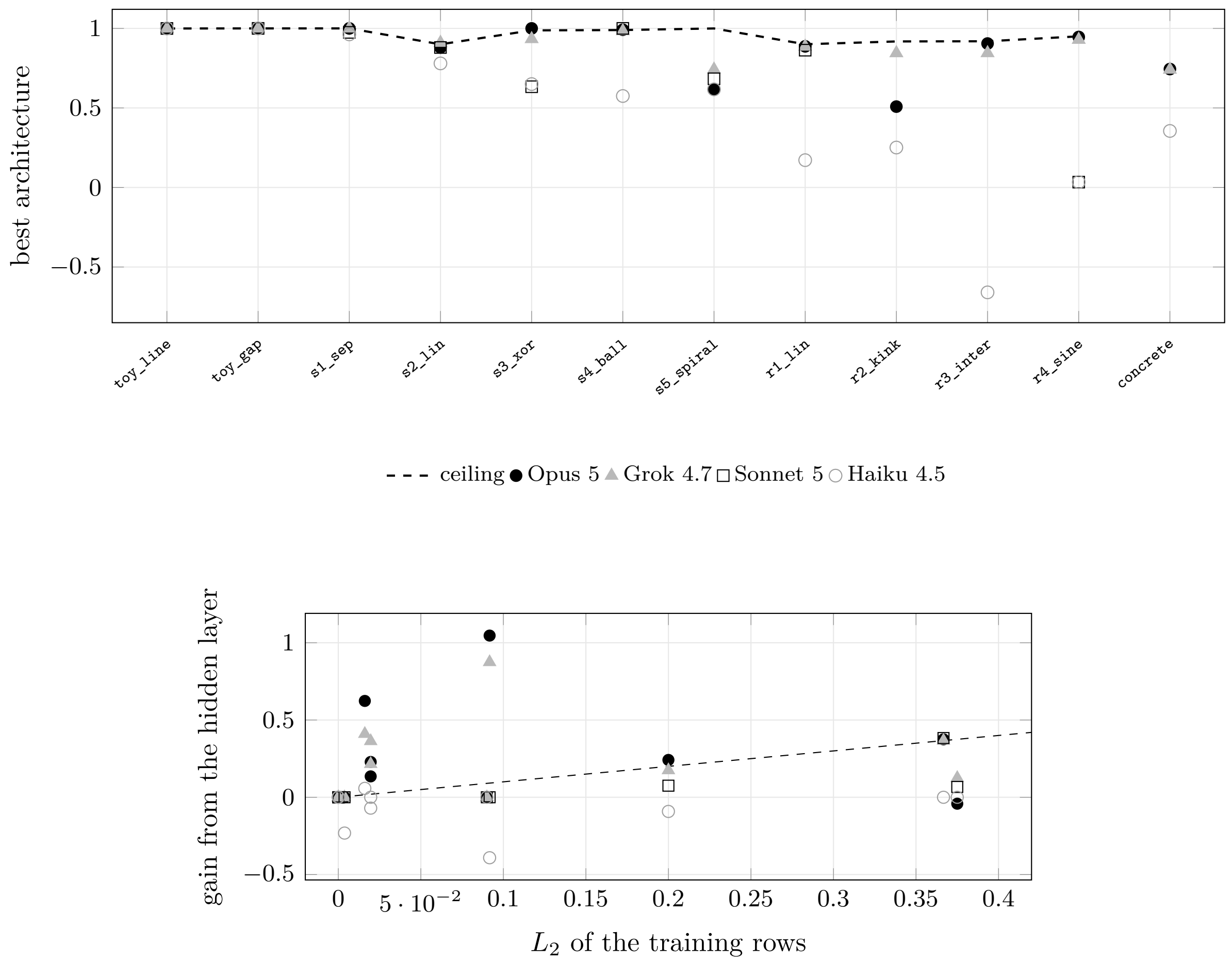
\begin{figure}[t]
\centering
\begin{tikzpicture}
\begin{axis}[
  name=top, width=0.96\textwidth, height=5.6cm,
  ylabel={best architecture}, ymin=-0.85, ymax=1.12,
  xtick={0,...,11}, xticklabels={\texttt{toy\_line},\texttt{toy\_gap},\texttt{s1\_sep},
    \texttt{s2\_lin},\texttt{s3\_xor},\texttt{s4\_ball},\texttt{s5\_spiral},\texttt{r1\_lin},
    \texttt{r2\_kink},\texttt{r3\_inter},\texttt{r4\_sine},\texttt{concrete}},
  xticklabel style={rotate=40, anchor=north east, font=\tiny},
  tick label style={font=\footnotesize}, label style={font=\small},
  grid=both, grid style={gray!18}, xmin=-0.6, xmax=11.6,
  legend style={font=\scriptsize, at={(0.5,-0.42)}, anchor=north, legend columns=5,
    draw=none, fill=none},
]
\addplot[black, thick, dashed, mark=none] table[x=i,y=ceiling] {figdata/arms_by_rung.dat};
\addlegendentry{ceiling}
\addplot[only marks, mark=*, mark size=2.1pt, black] table[x=i,y=opus5] {figdata/arms_by_rung.dat};
\addlegendentry{Opus 5}
\addplot[only marks, mark=triangle*, mark size=2.6pt, gray!55] table[x=i,y=grok] {figdata/arms_by_rung.dat};
\addlegendentry{Grok 4.7}
\addplot[only marks, mark=square, mark size=2.1pt, black] table[x=i,y=sonnet5] {figdata/arms_by_rung.dat};
\addlegendentry{Sonnet 5}
\addplot[only marks, mark=o, mark size=2.3pt, gray!75] table[x=i,y=haiku45] {figdata/arms_by_rung.dat};
\addlegendentry{Haiku 4.5}
\end{axis}
\begin{axis}[
  name=bot, at={(top.below south)}, anchor=north, yshift=-1.5cm,
  width=0.66\textwidth, height=5.0cm,
  xlabel={$L_2$ of the training rows}, ylabel={gain from the hidden layer},
  tick label style={font=\footnotesize}, label style={font=\small},
  grid=both, grid style={gray!18}, xmin=-0.02, xmax=0.42,
]
\addplot[domain=0:0.42, thin, dashed, black, forget plot] {x};
\addplot[only marks, mark=*, mark size=2.0pt, black] table[x=l2,y=gain] {figdata/arms_gain_opus5.dat};
\addplot[only marks, mark=triangle*, mark size=2.5pt, gray!55] table[x=l2,y=gain] {figdata/arms_gain_grok.dat};
\addplot[only marks, mark=square, mark size=2.0pt, black] table[x=l2,y=gain] {figdata/arms_gain_sonnet5.dat};
\addplot[only marks, mark=o, mark size=2.2pt, gray!75] table[x=l2,y=gain] {figdata/arms_gain_haiku45.dat};
\end{axis}
\end{tikzpicture}
\caption{Above, the four arms on every rung, each point the better of that arm's two architectures, with the ceiling of the generating process drawn as a dashed line and the real rung carrying none. The arms are indistinguishable on the left of the figure, where the answer is a line and the ceiling is nearly one, and they fan out from the parity rung onwards. Sonnet is absent on three rungs because it returned no parameter vector there across four dispatches. Haiku is the only arm that falls below zero, which on a regression rung means a network worse than predicting the mean of the training targets, and it does so on two rungs while predicting beforehand that it would score above $0.55$ on both. Below, what the hidden layer bought each arm against how far that rung's training rows are from linearly separable, with the identity drawn for reference. The relationship reported for one arm earlier in this paper, that the gain tracks $L_2$ where the construction succeeds, is a property of the stronger constructors rather than of the rungs: the points near the diagonal belong to the two arms that decline the hidden layer on exactly the five rungs where it is useless, and the points far below it belong to the arm that spends it where the optimum is linear.}
\Description{Two panels. The upper panel is a scatter plot with the twelve rungs along the horizontal axis and the held-out score of the better architecture on the vertical axis, with a dashed line marking the ceiling of each generating process; four series of markers, one per constructor arm, lie together near the ceiling on the leftmost four rungs and separate on the remaining eight, one series dropping below zero on two regression rungs and another absent on three. The lower panel plots the gain from adding a hidden layer against the training error of a linear separator, with the identity line drawn; the points of the two strongest arms cluster near that line while the weakest arm's points scatter well below it, including negative gains.}
\label{fig:arms}
\end{figure}

Figure~\ref{fig:arms} draws the whole comparison and relates what each arm's hidden layer bought it to how far the rows of that rung are from linearly separable.

\subsection{An open-weight size series, and where the capability runs out}
\label{sec:openweight}

The four arms above are four points, and all four come from models large enough that the question of where this capability begins is left untouched. We therefore administered the suite to a graded series of five open-weight models served locally on one laptop graphics card, from four hundred and ninety-four million parameters to eight billion, every one of them four-bit quantised by the same scheme so that size is the only quantity that varies across the series. The arms, the serving conditions, the order of administration and the rule that ends an arm's ascent were all written down before the first dispatch, in the pre-registration released with this paper, and the rule is the one the rest of this section already applies: the rungs are administered easiest first, and the next rung is administered only if both architectures of the current one are admissible under the unchanged validator and both score above the rung's trivial predictor. The first rung that fails ends that arm, and every rung above it is reported as not dispatched rather than as a missing value.

No arm reached a second rung. All five stopped on \texttt{toy\_line}, which asks for a line through twenty rows of one feature and on which the four arms of Table~\ref{tab:arms} score between $0.9996$ and $0.9997$ against a ceiling of $0.9994$. Table~\ref{tab:openweight-tabular} gives each arm's dispatch and the reason it stopped.

What the series shows is that the manner of failure is ordered by size while the outcome is not. The smallest arm never produced a module at all: it exhausted a budget of eight thousand tokens and then a doubled budget of sixteen thousand, in both cases by emitting the same import statement several thousand times in succession, and the reply contains none of the delimiters the brief asks for. The next two arms stopped of their own accord with most of their budget unspent, three hundred and ninety-three tokens in one case and six hundred and forty-nine in the other, and wrote something that looks like a submission and is not one: the one-and-a-half-billion arm returned modules that reference undefined names and compute their parameters with an arithmetic expression over a library call instead of stating them, and the three-billion arm returned modules that import a module which does not exist in the package it names and leave the second layer's weight and bias set to the constant \texttt{None}. The four-billion arm is the one whose stop is partly an artefact of the machine rather than of the model, and the detail matters enough to state: it exhausted its ceiling inside its own deliberation, having written the first of the two architectures and not the second, and the doubled ceiling the protocol prescribes for that case needs a context window this card cannot hold for a model of its size. The module it did finish settles the question that would otherwise hang over that arm, because that module is inadmissible on its own terms, declaring a weight matrix of shape $(1,)$ where the rung's encoding requires a two-dimensional one. The largest arm is the only one to produce a module the validator admits and the harness can score, and it scores $0.3245$ where the trivial predictor scores $-0.0171$ and every arm of Table~\ref{tab:arms} scores above $0.999$; its second architecture declares a bias of width eight against a weight matrix of shape $(1,8)$, so the rung is not delivered and the ascent ends there as the rule requires.

Three of the five arms therefore failed with the reply budget barely touched, which rules out the output ceiling as the binding constraint for them in the way that it was ruled out for the weaker of the four arms above. The failure these models share is narrower and more specific than a general weakness at the task. What each of them writes is a program that would compute a network, or a reference to one, or a sketch of one, where the brief asks for the numbers themselves, and the validator's syntax-tree restriction catches exactly that substitution. The capability this paper measures therefore has a floor somewhere above eight billion four-bit parameters, and that floor sits at the point where the artefact stops being a construction at all, well above the point where it would start predicting badly.

\begin{table}[t]
\caption{The open-weight size series on \texttt{toy\_line}, the easiest rung of the suite, with the smallest model first. Every arm is four-bit quantised and served locally under the conditions fixed before the first dispatch, and no arm reached a second rung. \enquote{Ceilings} lists the reply budgets the arm was given in order, the second present only where the server ended the first for length; \enquote{stop} is the server's own stop reason for each; \enquote{tokens} is the longest reply any dispatch produced. \enquote{Modules} counts the two architectures the rung asks for that arrived at all, \enquote{admissible} is the validator's verdict on them unchanged, and the last column is the better architecture's held-out coefficient of determination where one could be computed, against a floor of $-0.0171$ for the predictor that returns the training mean.}
\label{tab:openweight-tabular}
\small
\begin{tabular}{lrllrrlr}
\toprule
arm & $10^9$ par. & ceilings & stop & tokens & modules & admissible & $R^2$ \\
\midrule
\texttt{qwen25\_05b} & 0.49 & 8k/16k & length/length & 16384 & 0/2 & no & --- \\
\texttt{qwen25\_15b} & 1.54 & 8k & stop & 649 & 2/2 & no & --- \\
\texttt{qwen25\_3b} & 3.09 & 8k & stop & 393 & 2/2 & no & --- \\
\texttt{qwen3\_4b} & 4.02 & 8k & length & 8192 & 1/2 & no & --- \\
\texttt{ministral8b} & 8.02 & 8k & stop & 799 & 2/2 & no & $0.3245$ \\
\bottomrule
\end{tabular}
\end{table}

\section{Discussion}
\label{sec:discussion}

Section~\ref{sec:experiments} reported twelve rungs, four constructors and an
open-weight series. This section states what that collection supports, in four
steps: what the suite says about the shape of the capability, why the complexity
index fails to order it, what the two failures have in common, and how much of
that is a property of the practice rather than of one constructor. We close with
the limitations the design makes unavoidable.

The clearest result of the suite is that the capability is not well described by a single difficulty axis. On the toy and low-complexity rungs the hand-written network is indistinguishable from the convex optimum of the objective it is approximating. On the middle rungs it is either at the irreducible error or far from it, and which of the two happens is decided by whether the structure of the problem has a name the model can reach for. A parity rule has one, and the construction is then an identity rather than an approximation. A radial boundary has one, and eight rectified units make an octagon whose corners their own summation rounds off. A sinusoid has one, and the question becomes where to put eight kinks, which the model answered by deriving the optimal trapezoid rather than taking the obvious triangle. A spiral of varying pitch has no such name within the vocabulary of a fixed number of parallel half-planes, and that is the rung where the construction did worse than its own linear arm.

The second result is that the composite complexity index does not order the rungs the way the architecture gap does, and the disagreement is informative rather than a defect of the measurement. The index used here averages the error of a linear separator, the fraction of boundary points in a minimum spanning tree and the leave-one-out error of the nearest-neighbour rule, and by that average the parity rung is easier than the overlapping-Gaussian rung, at $0.086$ against $0.193$. It is easier, for a local learner: its nearest-neighbour error is $0.008$ and its boundary fraction is $0.050$, because a checkerboard of compact clusters is trivially solved by looking at what is nearby. It is harder for anything that has to write down a global rule, and the single measure that tracks what the hidden layer is worth is the error of the linear separator taken alone, which Figure~\ref{fig:gap} shows agreeing with the measured gain to within a few points on four of the five classification rungs. A practitioner choosing between architectures on the evidence of a complexity profile should therefore read its members rather than their average, and the member to read is the linearity group.

The third result concerns what the failures have in common. Both of them, and the one substantial failure of the benchmark tier, are failures to see structure that is oblique to the coordinates. The hinge rung was analysed one feature at a time and its folds run across both; the correlated synthetic rung of the benchmark tier was analysed under an assumption of diagonal covariance and its signal lies in the off-diagonal; the spiral was analysed in a single tilted coordinate whose tilt is correct only in part of the plane. Where the construction did handle an oblique structure it did so by first recovering a frame: the parity rung by locating four cluster centres and taking the sides of the parallelogram they form, the saddle by diagonalising the quadratic form it had estimated. That is a procedure the model clearly has, and the three failures are cases in which it did not reach for it.

The comparison in the preceding subsection settles part of that reading and leaves the rest open. If these were failures of the practice, every constructor shown the same rows would meet the same wall. One did not. On the hinge rung a second constructor reached $0.8448$ where the first reached $0.5084$, a difference that survives correction for multiple comparisons with a bootstrap interval on the difference in mean squared error of $[0.592, 1.365]$, and it did so by recovering the fold directions rather than by banding on one feature at a time. The oblique case is therefore within reach of a hand construction, and the failure on that rung belongs to the procedure the first constructor chose, which clears both the architecture it was filling in and the vocabulary available to it. The spiral points the same way without establishing it, since the second constructor is ahead there by $0.1667$ at an uncorrected $p$ of $0.0045$ that the Holm correction does not retain. The claim that survives is narrower than the one this paper would have made from a single arm, and it is also stronger, because it is now a statement about what a procedure can be rather than about what a model happened to do.

The limitations of the study are the ones its design makes unavoidable. There is one partition per rung, so the binomial noise on a classification score of a hundred and twenty held-out rows is about four and a half points at ninety-five per cent confidence, and small differences in the tables should not be read. The hand-written constructions are deterministic and cannot be averaged over seeds, which is why every comparison is reported paired on the same rows rather than as a difference of means. The main suite carries one constructing model, Claude Opus 5 at the \texttt{xhigh} reasoning effort the session log records, one pass per rung. Three further arms join it in the comparison, which makes four points on a curve this study does not trace; four models from two families, each at one effort setting, do not identify how the capability varies with the scale of the constructor. The open-weight series of Section~\ref{sec:openweight} adds five points below them and bounds the capability from underneath rather than extending the comparison, since none of its arms delivered a rung and the quantity it measures is therefore how each one failed rather than how well it scored. The gap between eight billion open-weight parameters and the frontier models of Table~\ref{tab:arms} is unsampled, so the location of the transition within that gap is not identified by anything reported here.

Three further limitations concern how the arms and the rungs were obtained. The arms differ in how they were obtained, in ways the dispatch log records per rung: one of them covers nine rungs rather than twelve, three of those nine carried a brevity instruction the others did not, and the four arms were not equally isolated from the sealed records. The last of these is the asymmetry that would most easily manufacture the observed ordering, and the only reason it does not explain it is that the most strictly isolated arm scores highest. No claim is made about any model or effort setting outside the four reported here. The synthetic rungs were generated by the same model that wrote the constructions, which knows the qualitative family of each one; the rotation, scaling and label permutation applied to every rung are the control for that, and they are a control on the numerical parameters and not on the family. The real rung is a single dataset, which is enough to show that the method transfers outside the laboratory and not enough to characterise how it behaves across real tabular problems in general.

\section{Conclusions}
\label{sec:conclusion}

A language model shown a small table of labelled rows and the shapes of a feed-forward network can write out every parameter of that network by reasoning alone, and the resulting frozen forward pass is, over a suite that runs from a twenty-row line fit to a thousand-mixture engineering regression, a competitive predictor. On the rungs whose optimum is linear it matches the closed-form optimum to within the noise of one partition. On three rungs whose optimum is not linear it reaches the irreducible error of the generating process, and on two of those it beats every fitted baseline including the same architecture trained by gradient descent. On two rungs it fails, in both cases because the structure it needed to see is oblique to the coordinates it analysed, and in both cases it had named the mechanism of its own failure in writing before the result was known.

The ordering of the suite by measured data complexity makes one further point. The quantity that predicts whether the hidden layer will earn its keep is how far the rows are from linearly separable, which the overall difficulty of the problem fails to track, and the two notions come apart sharply on exactly the rungs where this practice is most interesting.

What bounds these conclusions is set out in Section~\ref{sec:discussion}: one
partition per rung, deterministic constructions that cannot be averaged over
seeds, four constructors rather than a curve, a synthetic tier generated by the
same model that wrote the constructions, and a real tier of exactly one dataset.

Three directions follow. The first is to vary the constructing model and the
reasoning effort against a fixed suite, which would turn the present measurement
into the capability curve its arms are points on. The four arms reported here are
the first steps along that curve, and they already show that the capability does
not degrade along one dimension. All four are indistinguishable wherever the
answer is a line. Below that, three separate failures appear, and they are not
ordered versions of each other. A constructor may decline to deliver anything at
all on the hardest rungs while scoring well on the rest. It may deliver
everywhere and write networks worse than predicting the mean, while predicting
beforehand that they will be good. Or it may fail to produce a syntactically
admissible module on any rung. A curve drawn through scores alone would record
only the second of those three, and the first thing to go as the constructor
weakens is whether a well-formed artefact arrives at all.

The second direction is to give the construction a vocabulary for the oblique
case. All three failures recorded here are failures of frame rather than of
arithmetic, and the model demonstrably possesses the frame-recovery procedure it
did not apply, since it applied that same procedure on the rungs it solved.

The third direction, and the one this suite was built to make possible, is to ask
what complexity of model is required to perform weight-space programming at what
complexity of dataset. The rungs here carry measured complexity scores on a
standard axis, the constructions carry sealed digests and declared predictions,
and the protocol is cheap enough to repeat across a grid of models and effort
settings. The open-weight series reported here is the first column of that grid,
and it answers the question at the grid's easiest rung. Five models spanning four
hundred and ninety-four million to eight billion four-bit parameters were
administered a twenty-row line fit under a rule fixed in advance, and none
delivered it. The quantity the series separates them on is therefore how far into
the deliverable each one gets before it stops being a construction, which is
ordered by size where the outcome is not. Filling the rest of that grid, in
particular the unsampled region between eight billion open-weight parameters and
the frontier models that solve every rung, would turn weight-space programming
from a demonstrated capability into a measured one.

\printbibliography

\end{document}